\documentclass[10pt,twocolumn,letterpaper]{article}
\usepackage{iccv}

\usepackage{graphicx}
\usepackage{amsmath}
\usepackage{amssymb}
\usepackage{booktabs}

\usepackage{color}
\usepackage{multirow}
\usepackage{enumitem}
\setlist[enumerate]{leftmargin=*}

\usepackage[utf8]{inputenc} 
\usepackage[T1]{fontenc}    
\usepackage{url}            
\usepackage{booktabs}       
\usepackage{amsfonts}       
\usepackage{nicefrac}       
\usepackage{microtype}      
\usepackage{xcolor}         
\usepackage{ulem}
\usepackage{dsfont}
\usepackage{tabularx}
\usepackage{subcaption}

\definecolor{cvprblue}{rgb}{0.21,0.49,0.74}
\usepackage[pagebackref,breaklinks,colorlinks,allcolors=cvprblue]{hyperref}

\newcommand\blfootnote[1]{%
\begingroup
\renewcommand\thefootnote{}\footnote{#1}%
\addtocounter{footnote}{-1}%
\endgroup
}

\usepackage[capitalize]{cleveref}
\crefname{section}{Sec.}{Secs.}
\Crefname{section}{Section}{Sections}
\Crefname{table}{Table}{Tables}
\crefname{table}{Tab.}{Tabs.}

\def\paperID{******} 
\def\confName{ICCV}
\def\confYear{2025}

\begin{document}

\title{Improving Multi-Subject Consistency in Open-Domain Image Generation \\
with Isolation and Reposition Attention}

\author{
Huiguo He\textsuperscript{1,2}\thanks{This work was performed when Huiguo He (hehg3@mail2.sysu.edu.cn) was visiting 01.AI as a research intern.},
Qiuyue Wang\textsuperscript{2}, 
Yuan Zhou\textsuperscript{2}, 
Yuxuan Cai\textsuperscript{2},
Hongyang Chao\textsuperscript{1}, 
Jian Yin\textsuperscript{1$\dagger$},
Huan Yang\textsuperscript{2$\dagger$} \\
\textsuperscript{1}Sun Yat-Sun University, \textsuperscript{2}01.AI
}

\maketitle

\blfootnote{$\dagger$Huan Yang (hyang@fastmail.com) and Jian Yin are corresponding authors.}

\begin{abstract}
Training-free diffusion models have achieved remarkable progress in generating multi-subject consistent images within open-domain scenarios. The key idea of these methods is to incorporate reference subject information within the attention layer.  However, existing methods still obtain suboptimal performance when handling numerous subjects. This paper reveals two primary issues contributing to this deficiency. Firstly, the undesired internal attraction between different subjects within the target image can lead to the convergence of multiple subjects into a single entity.
Secondly, tokens tend to reference nearby tokens, which reduces the effectiveness of the attention mechanism when there is a significant positional difference between subjects in reference and target images.  To address these issues, we propose a training-free diffusion model with Isolation and Reposition Attention, named \textbf{IR-Diffusion}.
Specifically, \textbf{Isolation Attention} ensures that multiple subjects in the target image do not reference each other, effectively eliminating the subject convergence. On the other hand, \textbf{Reposition Attention} involves scaling and repositioning subjects in both reference and target images to the same position within the images. This ensures that subjects in the target image can better reference those in the reference image, thereby maintaining better consistency. Extensive experiments demonstrate that IR-Diffusion significantly enhances multi-subject consistency, outperforming all existing methods in open-domain scenarios.
\end{abstract}
\vspace{-15pt}

\section{Introduction}\label{sec:intro}

Multi-subject consistent image generation aims to produce visually engaging sequences that maintain consistency across multiple subjects, each based on its specific description. This is crucial for applications such as story visualization~\cite{SVSurvey} and game generation~\cite{videoGame1}. Recently, training-free diffusion models~\cite{storydiffusion, tewel2024_ConsiStory, dreamstory} have gained substantial attention for their ability to generate visually coherent frames. These models are especially suited to open-domain scenarios since they avoid the need for additional tuning or costly, high-quality, large-scale datasets.

\begin{figure}[t]
    \centering
    \includegraphics[width=1.0\linewidth]{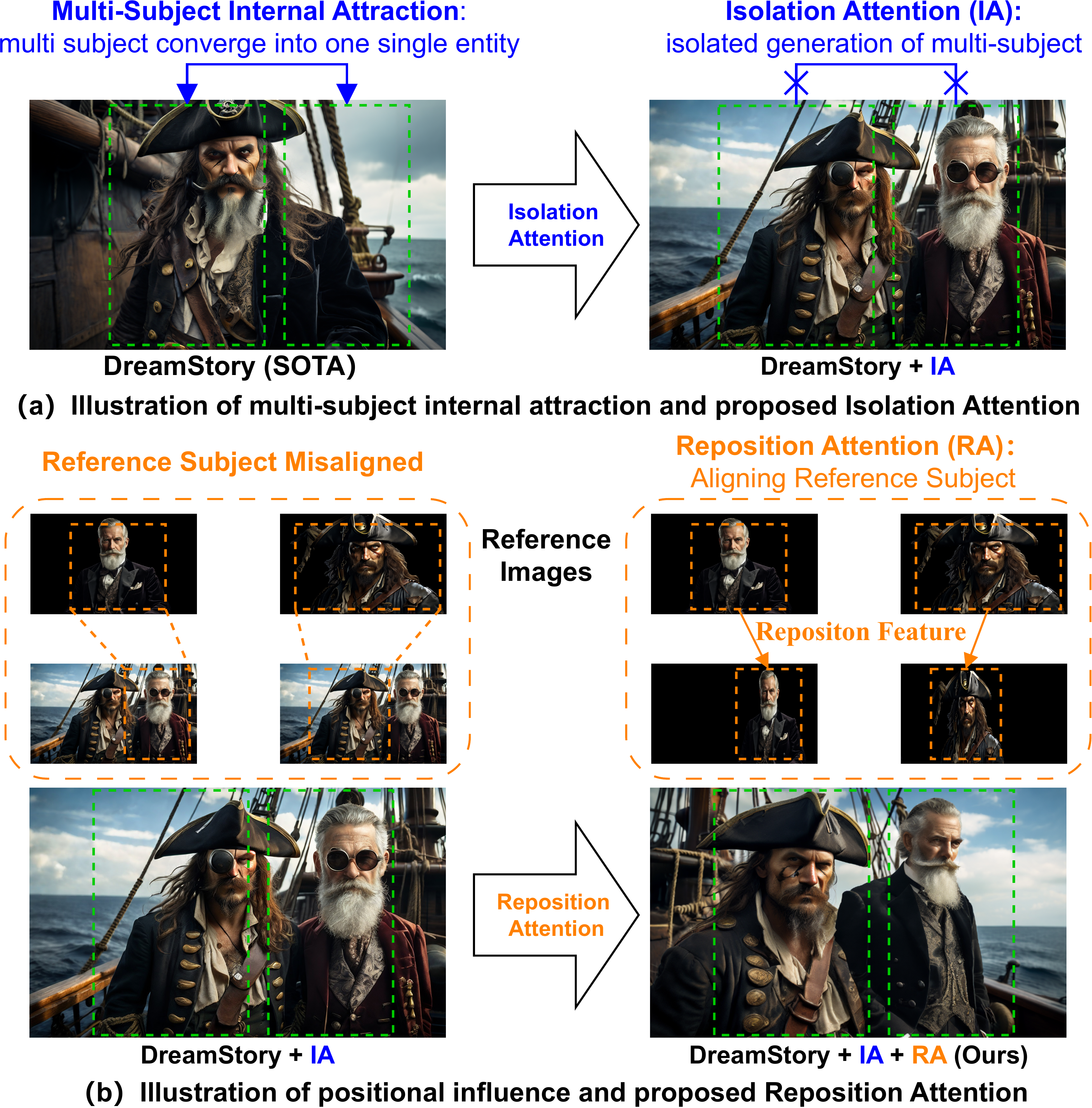}

    \caption{
    Illustration of our idea.
    (a) Internal attraction among subjects in the target image leads to the convergence of multiple subjects into a single entity. Our Isolation Attention (IA) effectively mitigates this problem, ensuring each subject can be independently generated.
    (b) Misalignment between subjects in the reference and target images results in the ineffective utilization of reference image features. Our Reposition Attention (RA) aligns the subjects in the reference and target images, thereby enabling the better utilization of the reference image and preserving fine-grained consistency.
    }
    \label{fig:finding}
    \vspace{-10pt}
\end{figure}

Training-free approaches for achieving subject consistency typically modify the attention mechanism to incorporate essential features from reference images and text. For instance, previous studies have primarily focused on incorporating features from reference images by substituting~\cite{cao2023masactrl} or concatenating~\cite{tewel2024_ConsiStory} the K and V in the self-attention layer with those from the reference images.
To address multi-subject consistency, DreamStory~\cite{dreamstory} introduces an innovative approach by using reference images and text as multimodal anchors (guidance). This method incorporates reference images into self-attention and text into cross-attention. 
A masking mechanism ensures that each subject aligns only with its \underline{external} respective anchor, mitigating subject fusion (blending).
This approach achieves SOTA multi-subject consistency in training-free diffusion models.
Nevertheless, these methods still fall short of optimal performance as they overlook inherent aspects of the attention mechanism within diffusion models, namely, multi-subject \underline{internal} attraction and position influences.

First, we observed there are undesired \underline{internal} attraction occurs among multiple subjects. This issue likely arises because diffusion models are typically trained on datasets containing a limited number of subjects, usually between 0 and 2. As a result, the model tends to generate fewer subjects, leading to possible subject convergence. As illustrated on the left of \cref{fig:finding}(a), the SOTA method DreamStory employs a masking mechanism to ensure that each subject can only reference its external corresponding reference. However, the internal attraction between subjects within the target image still results in a half-and-half merge, forming a composite character where the man constitutes the right half and the pirate the left half.
Second, tokens in the attention mechanism are inclined to focus on spatially proximate tokens. This tendency stems from a property learned by the diffusion model during pretraining, where it captures strong correlations between adjacent image pixels. As a result, the model assigns higher attention responses to nearby tokens, reducing the effectiveness of reference image features and leading to performance degradation when the subject positions in the reference and target images are misaligned. For example, on the left of \cref{fig:finding}(b), the man's clothing color is incorrect due to feature misalignment in the reference image.

To address these issues, we propose \textbf{IR-Diffusion}, a plug-and-play solution that incorporates Isolation and Reposition Attention. First, \textbf{Isolation Attention} (IA) minimizes internal attraction among multiple subjects by enforcing that each subject does not reference others within the target image during self-attention, preventing subject convergence, as shown on the right of \cref{fig:finding}(a). 
Secondly, \textbf{Reposition Attention} (RA) enhances subject alignment by rescaling and shifting reference images to match their positions in the target images. This can enable more effective use of reference images.  As shown on the right of \cref{fig:finding}(b), our IR-Diffusion (with both IA and RA) not only mitigates multi-subject convergence but also maintains detailed appearance consistency, leading to better multi-subject consistency in open-domain image generation.

The main contributions of this paper are as follows:
\vspace{-8pt}
\begin{enumerate}[itemsep=-3pt]
    \item To the best of our knowledge, we are the first to reveal and analyze the issues of multi-subject internal attraction and positional influence in the attention mechanism of diffusion models.
    \item We propose \textbf{Isolation Attention}, which prevents subject convergence and maintains the independence of each subject by minimizing internal attraction between them.
    \item We propose \textbf{Reposition Attention}, which ensures the use of reference information by aligning subjects to their corresponding positions.
    \item Extensive experiments demonstrate the advantages of our approach, \textbf{IR-Diffusion}. This training-free, plug-and-play solution is simple yet effective, and well-suited for open-domain scenarios. 
    Compared to the baseline DreamStory, IR-Diffusuion improved D\&C-DS metric by 0.19 (+75.4\%) on the DS-500 3-Subject benchmark.
\end{enumerate}

\section{Related Works}\label{sec:related_works}

Initially, Variational AutoEncoders (VAEs)~\cite{VAE} and Generative Adversarial Networks (GANs)~\cite{StoryGAN_2019_CVPR,song2020character_GAN, ma2022ai} are predominant in the field of image generation, but they inevitably face optimization challenges~\cite{ arjovsky2017WGAN, gulrajani2017improvedWGAN}. 
Subsequently, diffusion-based generative models~\cite{song2020DDIM_diffusion, DDPM_diffusion, song2020score_diffusion, nichol2021improved_diffusion, zhu2023moviefactory, wang2023videofactory, MVP} dominate this field. Notably, Stable Diffusion (SD)~\cite{sd_ldm_diffusion} applies diffusion techniques within latent space and is trained on the largest LAION-5B~\cite{schuhmann2022laion_5B} dataset, achieving impressive results and diversity in open-domain image generation. Despite these methods yielding promising results, they still struggle to guarantee subject consistency.

\subsection{Few-shot Finetuning Consistent Generation}

Few-shot finetuning methods~\cite{ruiz2023dreambooth, avrahami2023chosen, PortraitBooth, TPAMIcreateYourWorld, OneActor, multibooth, DCO, OMG, MUDI} finetune the model with a limited number of subject images to learn their unique textual expressions. For example, DreamBooth~\cite{ruiz2023dreambooth} introduces the concept of fine-tuning SD with LORA~\cite{hu2021lora} on several images to remember specific subject tokens. 
Kong et~al.~\cite{OMG} further train DreamBooth LoRA for every subject, then apply concept noise blending after a specific denoise timestep for multi-subject personalization. Jang et~al.~\cite{MUDI} propose using a segmentation model to separate subjects for training and inference, thus addressing the issue of multi-subject fusion. 
This method has achieved SOTA performance in few-shot finetuning for multi-subject consistent generation.

However, these approaches require finetuning for each individual story or subject, which incurs additional computational costs. Furthermore, this methodology inherently risks overfitting, potentially diminishing the aesthetic quality and diversity of the generated images~\cite{tewel2024_ConsiStory, dreamstory}.

\begin{figure*}[t]
    \centering
    \includegraphics[width=1.0\textwidth]{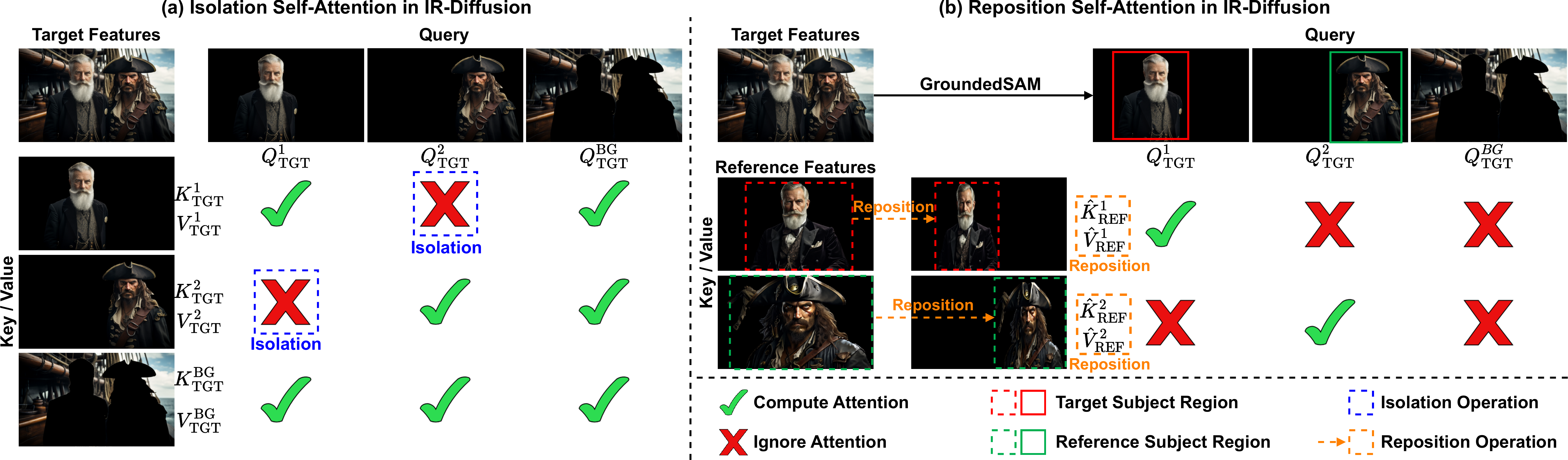}
    \caption{
        Illustration of our IR-Diffusion: 
        (a) Isolation Attention (IA): IA isolates internal attraction between different subjects by ensuring that subjects do not receive responses from the Key and Value of other subjects. 
        (b) Reposition Attention (RA): RA repositions the image features of the reference subjects to align with the positions of the corresponding subjects in the target image, enabling the model to more effectively utilize information from the reference image. 
        }
    \label{fig:framework}
    \vspace{-15pt}
\end{figure*}

\begin{figure}[t]
    \centering
    \setlength{\abovecaptionskip}{-0.001pt} 
    \includegraphics[width=1.0\columnwidth]{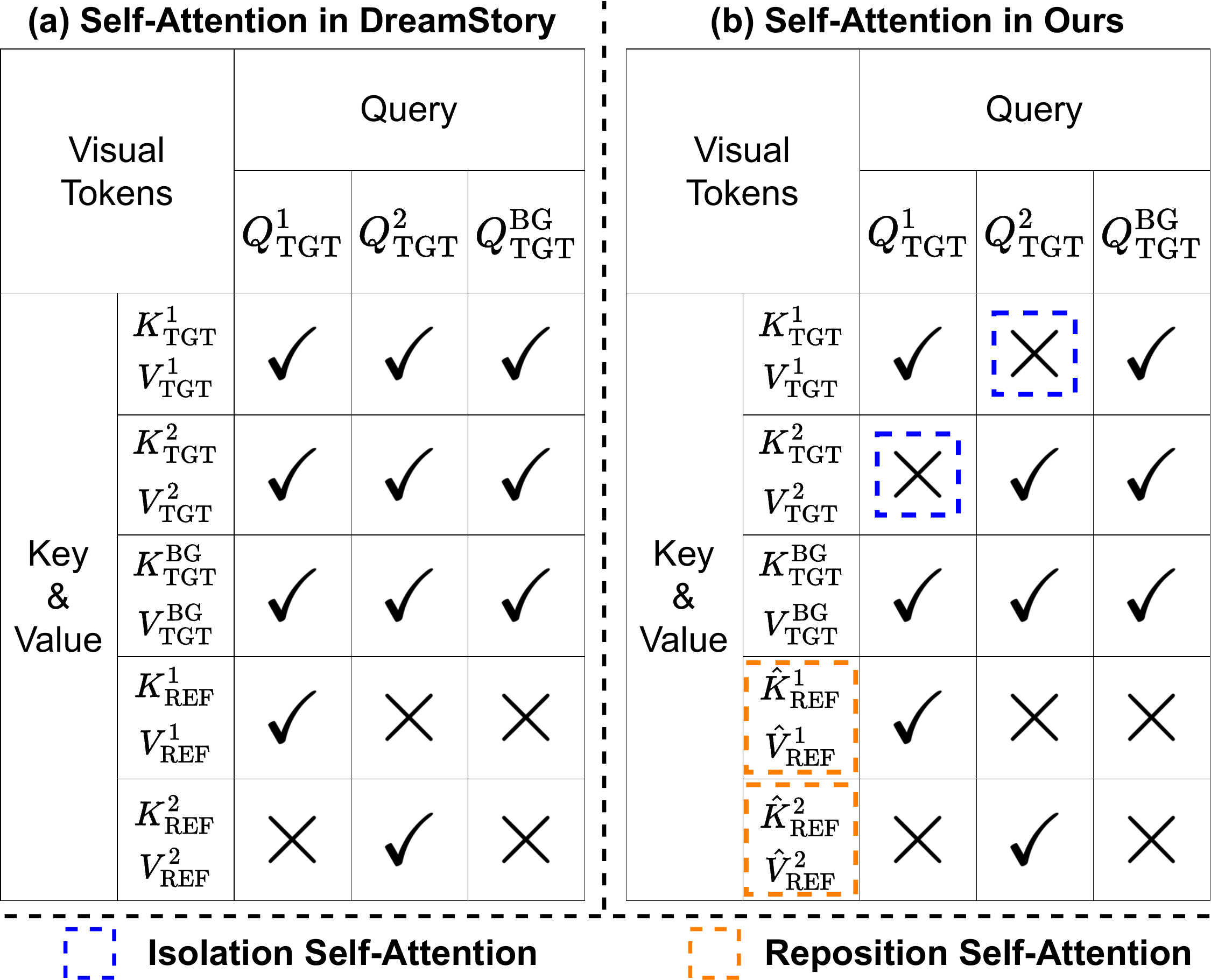}
    \caption{
        Comparison of the overall Self-Attention mechanism between DreamStory and our IR-Diffusion.
        This figure summarizes how the Queries, Keys, and Values are computed for different regions (subjects and background). The differences compared to the baseline (DreamStory~\cite{dreamstory}) are marked by colored dashed boxes: \textcolor{blue}{blue} represents IA, and \textcolor{orange}{orange} represents RA.
        }
    \label{fig:sa_compare}
    \vspace{-15pt}
\end{figure}

\subsection{Encoder-Based Consistent Generation}

In foundational T2I models like SD~\cite{sd_ldm_diffusion} and SDXL~\cite{podell2023sdxl}, cross-attention is crucial due to various conditions (e.g., text, semantic map) guiding the process through cross-attention~\cite{controlnet}. 
Previous methods~\cite{arar2023domain, IDAdapter, AnyDoor, PhotoMaker, ye2023_ip_adapter, PIAyanhong, BLIPdiffusion, ssr, MSDiffusion, InstantFamily, faceDiffusion, subjectDiffusion, JeDi} train image encoders for customized generation under image conditions. Specifically, some studies~\cite{IDAdapter, PhotoMaker, faceDiffusion, InstantFamily} train face encoders to maintain ID consistency by encoding subjects into image features and combining them with text features in cross-attention. Wang~et~al. propose MS-Diffusion~\cite{MSDiffusion}, which encodes images through a grounding resampler composed of several attention layers. MS-Diffusion then integrates multiple subjects in the cross-attention using a masking mechanism. They generate high-quality multi-subject datasets using their innovative data processing pipeline, achieving SOTA performance.

However, these methods are constrained by the scope of their datasets. They either support only human faces~\cite{IDAdapter, PhotoMaker, faceDiffusion, InstantFamily}, a limited range of domain~\cite{AnyDoor, ssr, MSDiffusion}, or a limited number of subjects, such as single-subject~\cite{ye2023_ip_adapter, BLIPdiffusion, JeDi} or two-subject~\cite{subjectDiffusion, storymaker}. Consequently, they are unsuitable for open-domain scenarios, where subjects can include humans, animals, and any number of subjects with various styles. 

\subsection{Training-free Consistent Generation}

Training-free methods~\cite{cao2023masactrl, tewel2024_ConsiStory, storydiffusion, theatergen, AutoStudio, OnePrompt, dreamstory} have garnered much attention because they do not require additional datasets or finetuning costs. The key idea of these methods is to enable the target image to reference the reference information within the attention layer. For instance, MasaCtrl~\cite{cao2023masactrl} introduced mutual self-attention, which substitutes the \textit{key} and \textit{value} in self-attention with those from the reference image. Later, ConsiStory~\cite{tewel2024_ConsiStory} and StoryDiffusion~\cite{storydiffusion} allow each frame to reference all subjects from multiple reference images within a batch. He et al.~\cite{dreamstory} propose the DreamStory framework, which uses LLM to parse story text and generate the necessary scene and subject information for consequent consistent scene generation. They introduce the innovative Masked Mutual Self-Attention (MMSA) and Masked Mutual Cross-Attention (MMCA), which enable the target image to leverage the corresponding reference subject's image and text, respectively. This multimodal referencing approach has achieved state-of-the-art results in open-domain multi-subject consistent generation.

However, such methods still yield sub-optimal performance because they overlook the intrinsic properties of the attention mechanism inherent to the diffusion model, i.e., the multi-subject internal attraction and influence of position.

\section{Our Approach}\label{sec:method}

The proposed \textbf{IR-Diffusion} is designed to generate multi-subject consistent images by leveraging reference subjects' texts and images, followed by the SOTA method DreamStory~\cite{dreamstory}. The process begins with generating individual subject images based on their textual descriptions. 
Subsequently, these generated images and their associated reference texts are utilized to produce a final composite image that maintains consistency across multiple subjects.

\subsection{Existing Attention Mechanism}

\begin{table}[t]
\setlength{\abovecaptionskip}{-0.01cm} 
\centering
\caption{Average response from other subjects and background (higher values are marked in bold). 
    The responses from other subjects are greater than those from the background, indicating internal attraction among subjects during the generation process.}
\label{tab:attraction}
\small
    \begin{tabular}{p{3.15cm} cc}
    \toprule[1.0pt]
    Average ($\times 10^{-3}$)                & Other Subjects & Background \\ \hline
    2-Subject &    \bf{0.1168}      &    0.1018   \\
    3-Subject &    \bf{0.1360}      &    0.1001   \\
    \toprule[1.0pt]
    \end{tabular}
    \vspace{-20pt}
\end{table}

In popular diffusion models (e.g., SD~\cite{sd_ldm_diffusion}, and SD-XL~\cite{podell2023sdxl}), the attention mechanism within the U-Net network typically consists of a self-attention layer followed by a cross-attention layer. The similarity between each token of image features is computed in self-attention to represent how one token responds to another, which is known as the attention map. All the responses are then normalized to $[0,1]$ by a softmax operation.
The attention layer can be formulated as follows, 
\begin{align}
\label{eq:attn}
    Attn(Q,K,V) &= softmax(QK) \cdot V,
\end{align}
Here, $Attn()$ represents the function of the attention layer, and $Q$, $K$, and $V$ represent the Query, Key, and Value features, respectively. These features are obtained by projecting the spatial features in the self-attention layers.

For clarity, we use the subscript ‘TGT’ to denote the features of the target image, and the subscript ‘REF’ to represent the features from the reference image. The superscript ‘\textit{i}’ indicates the \textit{i}-th subject. Additionally, in the target image, the region outside all subjects is defined as the background, denoted by the superscript ‘BG’. Thus, the features of the target image can be defined as follows:
\begin{align}
\label{eq:QKV_TGT}
    Q_\text{\scriptsize \tiny TGT} &= [Q_\text{\scriptsize \tiny TGT}^{1}, \ldots, Q_\text{\scriptsize \tiny TGT}^{N}, Q_\text{\scriptsize \tiny TGT}^{\text{\scriptsize \tiny BG}}],\\
    K_\text{\scriptsize \tiny TGT} & = [K_\text{\scriptsize \tiny TGT}^{1}, \ldots, K_\text{\scriptsize \tiny TGT}^{N}, K_\text{\scriptsize \tiny TGT}^{\text{\scriptsize \tiny BG}}],\\
    V_\text{\scriptsize \tiny TGT} & = [V_\text{\scriptsize \tiny TGT}^{1}, \ldots, V_\text{\scriptsize \tiny TGT}^{N}, V_\text{\scriptsize \tiny TGT}^{\text{\scriptsize \tiny BG}}],
\end{align}
where 'N' is the number of the subjects and $[\cdots]$ is the concatenation function.

Recent studies~\cite{tewel2024_ConsiStory, dreamstory,storydiffusion} have shown that appearance information can be incorporated into the self-attention layer by cascading the key ($K$) and value ($V$) features from the reference images. Based on these definitions, the output for the \textit{i}-th subject, $O^{i}$, can be calculated as follows,
\begin{align}
    O^{i} &= Attn(Q_{\text{\scriptsize \tiny TGT}}^{i},    [K_\text{\scriptsize \tiny REF}^{i}, K_\text{\scriptsize \tiny TGT}], 
    [V_\text{\scriptsize \tiny REF}^{i}, V_\text{\scriptsize \tiny TGT}]), \label{eq:KV_concat}
\end{align}
where the Key and Value of the \textit{i}-th reference subject are denoted as $K_\text{\scriptsize \tiny REF}^{i}$, and $V_\text{\scriptsize \tiny REF}^{i}$ respectively. In this way, the visual information of the reference subject is incorporated into the generation process to maintain consistent appearances. The background region does not require reference information from the reference subject. Thus, the output of the background, $O^{\text{\scriptsize \tiny BG}}$, can be expressed as the vanilla calculation:
\begin{align}
\label{eq:BG_vanilla}
    O^{\text{\scriptsize \tiny BG}} &= Attn(Q_\text{\scriptsize \tiny TGT}^{\text{\scriptsize \tiny BG}}, 
    K_\text{\scriptsize \tiny TGT}, V_\text{\scriptsize \tiny TGT}),
\end{align}

\subsection{Isolation Attention}\label{sec:iso_attn_sa}

\subsubsection{Internal Attraction in Self-Attention}\label{subsec:inter}

Diffusion pre-trained models~\cite{sd_ldm_diffusion, podell2023sdxl, li2024playground} are primarily designed for scenarios with a limited number of subjects, typically including only 1 or 2 target characters in their training samples. As a result, these models tend to generate a restricted number of characters.
When generating multiple subjects, the self-attention mechanism computes the similarity between all pairs of patches. This can lead to the undesired convergence of multiple similar characters into a single composite entity, as illustrated by the hybrid character of the pirate and the man in \cref{fig:finding}(a).

To investigate the attractions between multiple subjects in self-attention, we first generate an image with a fixed seed and use the GroundedSAM~\cite{GroundedSAM} to mark the positions of the subjects. All areas outside the subjects are defined as the background. We then regenerate the image using the same seed and calculate the response values. Specifically, we compute the average per-token response for each subject by dividing the total response from other subjects and the background by the number of tokens. The statistical results from the DS-500~\cite{dreamstory} benchmark are presented in \cref{tab:attraction}. As observed, the response from other subjects (0.1168) in the 2-subject scenario is higher than that of the background (0.1037). In the 3-subject scenario, this phenomenon is even more pronounced, with responses from other subjects at 0.1360 and background at 0.1001. These results all provide strong evidence of internal attraction among multiple subjects during the generation process.

\begin{figure}[t]
    \setlength{\abovecaptionskip}{-0.01cm} 
    \setlength{\belowcaptionskip}{-0.3cm}
    \centering

    \begin{subfigure}[t]{0.9\columnwidth}
        \centering
        \setlength{\abovecaptionskip}{-0.01cm}
        \includegraphics[width=\textwidth]{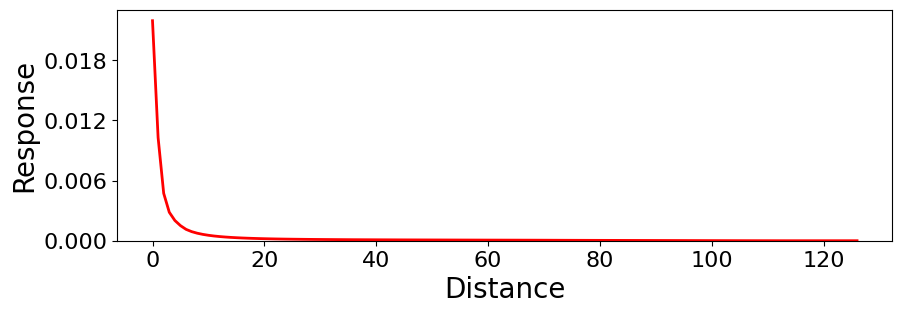}
        \caption{Relationship between Distance and Response (Scale 1)}
    \end{subfigure}
    
    \vspace{1em} 
    
    \begin{subfigure}[t]{0.9\columnwidth}
        \centering
        \setlength{\abovecaptionskip}{-0.01cm}
        \includegraphics[width=\columnwidth]{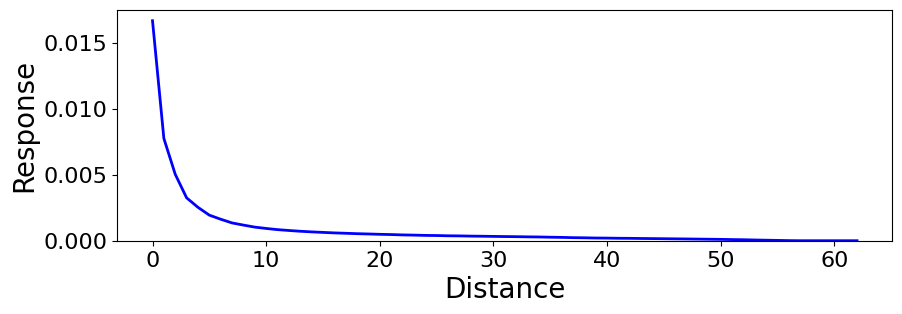}
        \caption{Relationship between Distance and Response (Scale 2)}
    \end{subfigure}

    \vspace{1em} 
    
    \caption{
    Average response values between tokens at varying distances in the self-attention layer.
    As the distance increases, the mean response values generally decrease, suggesting that tokens tend to reference nearby tokens more strongly.
    }
    \label{fig:position_curve}
    \vspace{-10pt}
\end{figure}

\begin{table*}[t]
    \setlength{\abovecaptionskip}{-0.1cm} 
    \centering
    \small \renewcommand{\arraystretch}{1.25}
    \caption{Quantitative comparison with other SOTA methods on the benchmark: Our RI-Diffusion demonstrates comparable performance on the aesthetic score (AES) and CLIP-T metrics while outperforming other methods on consistency metrics DC and D\&C-DS (with the best performance marked in bold).}
    \label{tab:results_benchmark}

\begin{tabular}{p{3.8cm}| c c c c | c c c c}
\bottomrule[0.8pt]
\multicolumn{1}{l|}{\multirow{2}{*}{}}                                                 & \multicolumn{4}{c|}{\textbf{2-Subject}} & \multicolumn{4}{c}{\textbf{3-Subject}} \\
\multicolumn{1}{l|}{}                                                                  & AES$\uparrow$ & CLIP-T$\uparrow$  & DS$\uparrow$  &  D\&C-DS$\uparrow$   & AES$\uparrow$ & CLIP-T$\uparrow$  & DS$\uparrow$  &  D\&C-DS$\uparrow$   \\ \hline
MuDI~\cite{MUDI}                                & 6.47    & 0.3652   & 0.6578     &  0.4410    &   6.54   &  0.3664   &  0.5924  &  0.1988  \\
MS-Diffusion~\cite{MSDiffusion}  & 6.12    & 0.3470   & 0.7386     &  0.6251    &   6.08   &  0.3502   &  0.6641  &  0.3617  \\
ConsiStory~\cite{tewel2024_ConsiStory}  & 6.62    & 0.3757   & 0.5988   & 0.4251  & 6.73   & 0.3770   & 0.5564   & 0.2038  \\
StoryDiffusion~\cite{storydiffusion}    & 6.56    & 0.3702   &      0.6258   &      0.4364  & 6.57   & 0.3707   & 0.5723   & 0.2095     \\
DreamStory~\cite{dreamstory}  & 6.72    & 0.3779   & 0.6714   &   0.5444 & 6.81  & 0.3791   & 0.5965   &  0.2335 \\
\textbf{IR-Diffusion} (\textbf{Ours})              & 6.64    & 0.3679   & \bf{0.7518}   &   \bf{0.6458} & 6.68   & 0.3736   & \bf{0.6742}   &  \bf{0.4095} \\ 
\toprule[0.8pt]
\end{tabular}
\vspace{-20pt}
\end{table*}

\subsubsection{Isolation Self-Attention}\label{subsubsec:inter}

As discussed in \cref{subsec:inter}, different subjects may be attracted to each other in the self-attention layer. This internal attraction can lead to subject convergence, where multiple subjects merge into a single entity, that simultaneously exhibits half characteristics of both original subjects. 

To mitigate this internal attraction, each subject should not affect the others during the vanilla forward process of the target image. For example, in \cref{fig:framework}, the subjects 'man' and 'pirate' should not attracted to each other. We achieve this by ensuring that each subject does not receive responses from the Key and Value of other subjects. 
By excluding other subjects' $K$ and $V$ from the original $K_\text{\scriptsize \tiny TGT}$ and $V_\text{\scriptsize \tiny TGT}$, the \cref{eq:KV_concat} can be modified as:
\begin{align}
\label{eq:KV_concat_new}
    O^{i} &= Attn(Q_{\text{\scriptsize \tiny TGT}}^{i}, 
    [K_\text{\scriptsize \tiny REF}^{i}, K_\text{\scriptsize \tiny TGT}^{i}, K_\text{\scriptsize \tiny TGT}^{\text{\scriptsize \tiny BG}}],
    [V_\text{\scriptsize \tiny REF}^{i}, V_\text{\scriptsize \tiny TGT}^{i}, V_\text{\scriptsize \tiny TGT}^{\text{\scriptsize \tiny BG}}]).
\end{align}
Compared to existing methods like DreamStory~\cite{dreamstory} (\cref{eq:KV_concat}), the key distinction of the proposed IA is that the $Q$ of subject $i$ in the target image does not reference the $KV$ of other subjects in the same image. This IA, highlighted by the blue dashed box in \cref{fig:sa_compare}, ensures that each subject is less attracted by other subjects, thereby preventing internal attraction from other subject regions in the target image.
As a result, each subject can be generated relatively independently, effectively avoiding the issue of subjects convergence. 

We apply Isolation Self-Attention in all decoder layers of the U-Net. This unconventional attention mechanism is implemented using a masking strategy, which zeroes out the undesired attention map responses between different subjects. More details on implementation can be found in the supplementary material.

\begin{figure*}[t]
    \centering
    \setlength{\abovecaptionskip}{-0.03cm}
    \includegraphics[width=0.95\textwidth]{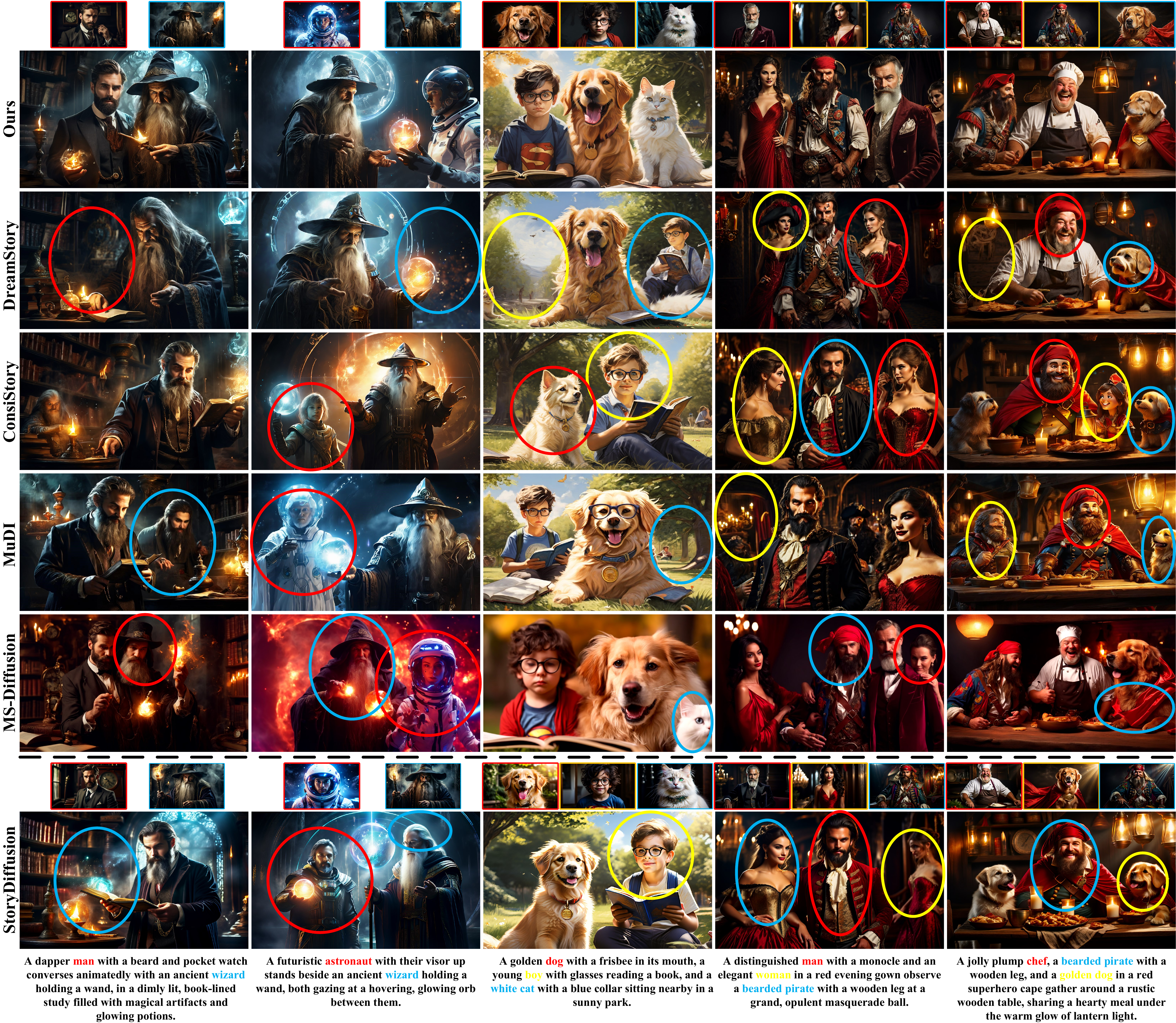}
    \caption{Comparisons of multi-subject consistency generation between our IR-Diffusion and other SOTA methods. The superior performance of our approach is evident from the more visually appealing and consistent results. Except for StoryDiffusion, which uses the portraits above it for reference, all other methods use the top portraits as a reference. Different subjects are indicated with different colors. 
    }
    \label{fig:main_results}
    \vspace{-15pt}
\end{figure*}

\subsection{Reposition Attention}\label{sec:relo_attn_sa}

\subsubsection{Positional Influence in Self-Attention}\label{subsec:loc}

Adjacent pixels generally exhibit strong correlations in typical images. Pre-trained on large image datasets, the diffusion model inherits this characteristic, prioritizing information from neighboring tokens. Additionally, the scale and crop conditions in the diffusion model~\cite{podell2023sdxl, li2024playground} might also contribute to this phenomenon, with these positional relationships being learned within the timestep embeddings.

Given a feature map of resolution $H \times W$ (comprising $HW$ tokens), the attention layer generates a corresponding $HW \times HW$ response matrix, referred to as the attention map. For any two tokens, let $\Delta X$ and $\Delta Y$ denote the respective horizontal and vertical positional offsets, We define the inter-token distance as the sum of these positional differences, i.e., $D = \left| \Delta X \right| + \left|\Delta Y\right|$. Since the U-Net architecture employs features across two distinct scales, we computed the mean responses from attention maps over varying resolutions and distances, which were aggregated across all steps and layers.
We exclude tokens within a 5\% margin along the edges of the image, as these tokens carry substantial positional information and are prone to noise~\cite{position1, position2, position3}.
We present the response curves for the two different scales under the DS-500~\cite{dreamstory} benchmark in \cref{fig:position_curve}. As shown in \cref{fig:position_curve}, the mean response values between tokens progressively decrease as the distance increases, indicating that tokens tend to reference nearby tokens more strongly within self-attention. 

\subsubsection{Reposition Self-Attention}\label{subsec:relo_attn_sa}

As discussed above, self-attention tends to prioritize tokens in neighboring positions. 
This reduces the utilization of information from the reference image when there is a substantial distance between the subjects in the reference image and their corresponding subjects in the target image.
One straightforward approach is to align the subjects to their corresponding positions in the self-attention layer. As shown in \cref{fig:framework}, the reference features (K and V) of the subject 'man' and 'pirate' will be rescaled and shifted to the position of the corresponding subject in the target image.

This repositioned feature will be denoted with a hat, i.e., $\hat{K}_\text{\scriptsize \tiny REF}$ and $\hat{V}_\text{\scriptsize \tiny REF}$. These features are rescaled and shifted before the projection to prevent disrupting the self-attention calculation process. Regions outside the repositioned reference subject will be filled with zeros to avoid introducing unnecessary noise. After applying Reposition Attention, the final attention operation in \cref{eq:KV_concat_new} can be formulated as follows:
\begin{align}
\label{eq_softmax}
    O^{i} &= Attn(Q_{\text{\scriptsize \tiny TGT}}^{i}, 
    [\hat{K}_\text{\scriptsize \tiny REF}^{i}, K_\text{\scriptsize \tiny TGT}^{i}, 
    K_\text{\scriptsize \tiny TGT}^{\text{\scriptsize \tiny BG}}], 
    [\hat{V}_\text{\scriptsize \tiny REF}^{i}, V_\text{\scriptsize \tiny TGT}^{i}, 
    V_\text{\scriptsize \tiny TGT}^{\text{\scriptsize \tiny BG}}]).
\end{align}
This strategy aligns the reference with the position of the target subject, thereby improving the efficiency of information utilization and preserving the fine-grained appearance consistency of the subjects. This design is similar to Positional Embeddings (PE) methods~\cite{RoPE_ViT, LaPE}, which enhance nearby token interactions by encoding spatial relationships. By enforcing positional alignment, RA similarly leverages spatial relationships to enhance information propagation.

Similar to IA, RA is implemented using the same masking strategy to eliminate undesired attention map responses between different components. For more details, please refer to the supplementary material.

\section{Experiments}\label{sec:exp}

\subsection{Implementation Details} \label{subsec:exp_implementation_details}

The Playground~\footnote{\href{https://huggingface.co/playgroundai/playground-v2.5-1024px-aesthetic}{playground-v2.5-1024px-aesthetic}} is adopted as our T2I backbone due to its superior performance in open-domain multi-subject consistent generation, which is also be used by DreamStory~\cite{dreamstory}. 
To obtain the subject mask for the target image, we follow DreamStory by performing a rehearsal generation with the original backbone to identify subject masks. 
More details can be found in the Supplementary.

\begin{table*}[t]
\setlength{\abovecaptionskip}{-0.1cm} 
    \centering
    \small \renewcommand{\arraystretch}{1.25}
    \caption{Quantitative results of the ablation study on the benchmark: Reposition Attention (RA) and Isolation Attention (IA) are individually incorporated into the baseline. The aesthetic score (AES) and CLIP-T metrics show minimal change, while a significant improvement is observed in the consistency metrics DC and D\&C-DS (with the best performance marked in bold).}
    \label{tab:results_ablation_benchmark}

\begin{tabular}{p{3.8cm}| c c c c | c c c c}
\bottomrule[0.8pt]
\multicolumn{1}{l|}{\multirow{2}{*}{}}                              & \multicolumn{4}{c|}{\textbf{2-Subject}} & \multicolumn{4}{c}{\textbf{3-Subject}} \\
\multicolumn{1}{l|}{}                                               & AES$\uparrow$ & CLIP-T$\uparrow$  & DS$\uparrow$  &  D\&C-DS$\uparrow$   & AES$\uparrow$ & CLIP-T$\uparrow$  & DS$\uparrow$  &  D\&C-DS$\uparrow$   \\ \hline

Baseline (DreamStory~\cite{dreamstory})  & 6.61    & 0.3681   & 0.6819   & 0.5592  & 6.77   & 0.3655   & 0.6042   & 0.2378   \\
\quad w/ RA          & 6.69   & 0.3668   & 0.7282   &  0.5978    & 6.80 & 0.3613   & 0.6323 & 0.2405 \\
\quad w/ IA          & 6.67   & 0.3679   & 0.7453   &  0.6364    & 6.76 & 0.3729   &0.6695 & 0.3636 \\
\quad w/ RA+IA (\textbf{Ours}) & 6.64 & 0.3679   & \bf{0.7518} & \bf{0.6458} & 6.68   & 0.3736   & \bf{0.6742}   &  \bf{0.4095}  \\ \toprule[0.8pt]
\end{tabular}
\vspace{-20pt}
\end{table*}

\subsection{Evaluation Benchmark} \label{subsec:exp_benchmark}

We adopt the DS-500 benchmark to evaluate the performance of our approach, following DreamStory~\cite{dreamstory}. This benchmark provides a comprehensive validation for multi-subject scenarios in open-domain multi-subject consistent image generation. Specifically, each subject's portrait is generated from its prompt using the same diffusion model, i.e., Playground~\cite{li2024playground} in our experiments. These portraits and prompts then serve as multimodal references for consistent multi-subject scene generation.

\subsection{Evaluation Metrics}  \label{subsec:exp_metrics}

We follow the evaluation metrics in DreamStory~\cite{dreamstory} to assess the performance of our method. Specifically, we evaluate generated results based on three main criteria: aesthetics, image-text alignment, and subject consistency. For objective evaluation, we adopt the predictor~\footnote{\href{https://github.com/christophschuhmann/improved-aesthetic-predictor}{improved-aesthetic-predictor}} for the aesthetic score (\textbf{AES}), CLIP~\footnote{\href{https://huggingface.co/openai/clip-vit-base-patch16}{clip-vit-base-patch16}} score for text-image alignment (\textbf{CLIP-T}), DreamSim~\cite{fu2024dreamsim} for single subject consistency (\textbf{DS}), and \textbf{D\&C-DS}~\cite{MUDI} to evaluate the multi-subject consistency.

To further validate the results, we conduct a user study using an A/B test. For each evaluation, we randomly display two sets of images accompanied by their respective texts. Each set is generated using a different method. Participants judge each metric by selecting whether Image A is superior, Image B is superior, or both are comparable. We collect 2000 votes for each comparison. The final results are aggregated and presented as percentages.

\begin{figure}[t]
    \centering
    \setlength{\abovecaptionskip}{-0.03cm}
    \includegraphics[width=1.0\columnwidth]{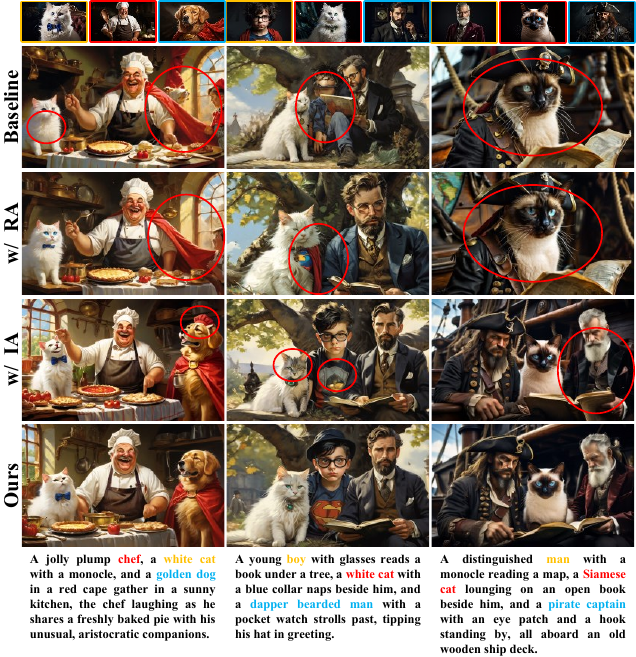}
    \caption{
    Ablation studies of different generation results.
    All methods use the top portraits as a reference. Different subjects are indicated with different colors. More ablation studies can be found in the supplementary materials.
    }
    \label{fig:ablation_study}
    \vspace{-15pt}
\end{figure}

\subsection{Comparison with SOTA Methods} \label{subsec:exp_compare_SOTA}

We conduct exhaustive experiments to compare our method with other SOTA methods, including the finetuning-based MuDi~\cite{MUDI}, the encoder-based MS-Diffusion~\cite{MSDiffusion}, and several training-free methods, i.e., StoryDiffusion~\cite{storydiffusion}, ConsiStory~\cite{tewel2024_ConsiStory}, and DreamStory~\cite{dreamstory}. 
For a fair comparison, MS-Diffusion also uses the same mask with proportional positioning for layout-to-image generation.

\subsubsection{Objective Comparison}

The quantitative comparison results with existing SOTA methods are presented in \cref{tab:results_benchmark}. Our IR-Diffusion approach achieves performance comparable to the best existing methods in terms of aesthetic score (AES) and image-text similarity (CLIP-T), and surpasses them in consistency metrics, specifically DC and D\&C-DS. 
Notably, compared to the current training-free SOTA method (DreamStory~\cite{dreamstory}), IR-Diffusion improves the D\&C-DS metric by approximately 0.10 (18.6\%) on the 2-Subject benchmark and 0.19 (75.4\%) on the 3-Subject benchmark.
Compared to the 2-subject scenario, the 3-subject scenario shows a more significant improvement. This is due to the higher likelihood of subject misalignment and internal attraction in the 3-subject scenario. Our method effectively addresses these challenges, leading to a more substantial performance enhancement.

Our approach also outperforms both MuDI~\cite{MUDI} and MS-Diffusion~\cite{MSDiffusion} across all four metrics. 
MS-Diffusion yields the lowest AES score (around 6.1), possibly due to the lower quality of images in their dataset, mainly from general videos.
Additionally, our method avoids the tuning costs and overfitting risks associated with MuDI, as well as MS-Diffusion’s limitations in supporting certain subject types due to constraints in their training dataset.
Overall, these results demonstrate the advancement of our method.

\begin{table*}[t]
    \centering
    \small 
    \setlength{\abovecaptionskip}{-0.03cm} 
    \caption{User study on aesthetic scores (AES), text-image relevance (T-I Align), and multi-subject consistency.
        Values are presented as percentages (with the \% symbol omitted). The winning dimension in the consistency is highlighted in \textbf{bold}.}
    \label{tab:userstudy_SOTA}


    \begin{tabular}{p{3.5cm}|cccccc|cccccc}
        \hline
         &
          \multicolumn{6}{c|}{2-Subject} &
          \multicolumn{6}{c}{3-Subject} \\ \cline{2-13} 
        \multirow{2}{*}{Dimensions} &
          \multicolumn{2}{c|}{AES} &
          \multicolumn{2}{c|}{T-I Align} &
          \multicolumn{2}{c|}{Consistency} &
          \multicolumn{2}{c|}{AES} &
          \multicolumn{2}{c|}{T-I Align} &
          \multicolumn{2}{c}{Consistency} \\
         &
          Win &
          \multicolumn{1}{c|}{Lose} &
          Win &
          \multicolumn{1}{c|}{Lose} &
          Win &
          Lose &
          Win &
          \multicolumn{1}{c|}{Lose} &
          Win &
          \multicolumn{1}{c|}{Lose} &
          Win &
          Lose \\ \hline
        \multicolumn{1}{l|}{DreamStory~\cite{dreamstory}} &
          11.9 &
          \multicolumn{1}{c|}{4.7} &
          21.4 &
          \multicolumn{1}{c|}{25.8} &
          \textbf{51.3} &
          \multicolumn{1}{c|}{6.9} &
          36.7 &
          \multicolumn{1}{c|}{6.4} &
          63.1 &
          \multicolumn{1}{c|}{13.3} &
          \textbf{66.6} &
          4.9 \\
        \multicolumn{1}{l|}{ConsiStory~\cite{tewel2024_ConsiStory}} &
          18.3 &
          \multicolumn{1}{c|}{7.3} &
           43.7 &
          \multicolumn{1}{c|}{15.0} &
          \textbf{80.7} &
          \multicolumn{1}{c|}{2.3} &
          10.3 &
          \multicolumn{1}{c|}{7.3} &
          27.0 &
          \multicolumn{1}{c|}{10.6} &
          \textbf{77.7} &
          1.6 \\
        \multicolumn{1}{l|}{StoryDiffusion~\cite{storydiffusion}} &
          10.0 &
          \multicolumn{1}{c|}{3.6} &
          28.3 &
          \multicolumn{1}{c|}{14.0} &
          \textbf{82.0} &
          \multicolumn{1}{c|}{3.3} &
          7.5 &
          \multicolumn{1}{c|}{6.0} &
          23.5 &
          \multicolumn{1}{c|}{15.0} &
          \textbf{79.0} &
          5.0 \\

          \multicolumn{1}{l|}{MuDI~\cite{MUDI}} &
          8.0 &
          \multicolumn{1}{c|}{5.0} &
          29.0 &
          \multicolumn{1}{c|}{3.0} &
          \textbf{65.0} &
          \multicolumn{1}{c|}{1.7} &
          19.5 &
          \multicolumn{1}{c|}{3.5} &
          27.0 &
          \multicolumn{1}{c|}{1.0} &
          \textbf{73.0} &
          2.5 \\

          \multicolumn{1}{l|}{MS-Diffusion~\cite{MSDiffusion}} &
          43.3 &
          \multicolumn{1}{c|}{3.7} &
          34.7 &
          \multicolumn{1}{c|}{4.3} &
          \textbf{51.7} &
          \multicolumn{1}{c|}{15.7} &
          44.8 &
          \multicolumn{1}{c|}{8.5} &
          43.5 &
          \multicolumn{1}{c|}{2.5} &
          \textbf{46.8} &
          17.7 \\
        \hline
        \end{tabular}
    
    \vspace{-0.3cm}
\end{table*}

\begin{table*}[t]
    \centering
    \small 
    \setlength{\abovecaptionskip}{-0.01cm} 

    \caption{User study on aesthetic scores (AES), text-image relevance (T-I Align), and multi-subject consistency. 
            Values are presented as percentages (with the \% symbol omitted). The winning dimension in the consistency is highlighted in \textbf{bold}.}
    \label{tab:userstudy_abl}

    \begin{tabular}{p{3.25cm}|cccccc|cccccc}
        \hline
         &
          \multicolumn{6}{c|}{2-Subject} &
          \multicolumn{6}{c}{3-Subject} \\ \cline{2-13} 
        \multirow{2}{*}{Dimensions} &
          \multicolumn{2}{c|}{AES} &
          \multicolumn{2}{c|}{T-I Align} &
          \multicolumn{2}{c|}{Consistency} &
          \multicolumn{2}{c|}{AES} &
          \multicolumn{2}{c|}{T-I Align} &
          \multicolumn{2}{c}{Consistency} \\
         &
          Win &
          \multicolumn{1}{c|}{Lose} &
          Win &
          \multicolumn{1}{c|}{Lose} &
          Win &
          Lose &
          Win &
          \multicolumn{1}{c|}{Lose} &
          Win &
          \multicolumn{1}{c|}{Lose} &
          Win &
          Lose \\ \hline
        \multicolumn{1}{l|}{Baseline (DreamStory~\cite{dreamstory})} &
          9.4 &
          \multicolumn{1}{c|}{3.6} &
          23.0 &
          \multicolumn{1}{c|}{32.0} &
          \textbf{52.2} &
          \multicolumn{1}{c|}{5.4} &
          33.5 &
          \multicolumn{1}{c|}{11.5} &
          54.0 &
          \multicolumn{1}{c|}{5.8} &
          \textbf{66.0} &
          5.5 \\
        \multicolumn{1}{l|}{\quad w/ RA } &
          1.7 &
          \multicolumn{1}{c|}{1.0} &
          8.0 &
          \multicolumn{1}{c|}{7.7} &
          \textbf{31.3} &
          \multicolumn{1}{c|}{6.7} &
          6.0 &
          \multicolumn{1}{c|}{11.0} &
          43.0 &
          \multicolumn{1}{c|}{7.2} &
          \textbf{40.0} &
          8.6 \\
        \multicolumn{1}{l|}{\quad w/ IA} &
          6.7 &
          \multicolumn{1}{c|}{3.3} &
          6.7 &
          \multicolumn{1}{c|}{3.3} &
          \textbf{7.7} &
          \multicolumn{1}{c|}{1.3} &
          23.3 &
          \multicolumn{1}{c|}{13.3} &
          5.0 &
          \multicolumn{1}{c|}{3.0} &
          \textbf{17.3} &
          6.0 \\
        \hline
        \end{tabular}

    \vspace{-15pt}
\end{table*}

\subsubsection{Subjective Comparison} 

\cref{tab:userstudy_SOTA} presents the results of our user study compared with SOTA methods. This table demonstrates that our approach surpasses all methods in open-domain scenarios regarding consistency. Notably, it demonstrates a greater advantage over MS-Diffusion in the user study compared to objective metrics. This discrepancy arises from the composition of MS-Diffusion’s dataset, which predominantly features clothing items rather than facial features. As a result, the model performs better on clothing, while users are more sensitive to the accuracy of facial features. More experimental results and discussions can be found in the supplementary material.

We also show the generated images of IR-Diffusion and other SOTA methods in \cref{fig:main_results}. As shown in \cref{fig:main_results}, our IR-Diffusion preserves multi-subject consistency, e.g., wizard and astronaut in the first and second columns, respectively. Specifically, MS-Diffusion struggles with certain subjects (e.g., wizards and astronauts) and often produces a red tint, possibly due to their limited dataset and the leaking of ground tokens.
These results highlight our method's superiority in generating multi-subject consistent images in open-domain scenarios, surpassing all previous methods.

\subsection{Ablation Studies} \label{subsec:exp_ablation}

We also conduct an ablation study to validate the effectiveness of each component in our method. Specifically, we add the Reposition Attention (RA) and the Isolation Attention (IA) modules individually. We evaluate DreamStory~\cite{dreamstory} under the same settings as our baseline. 

\subsubsection{Objective Comparison}

The quantitative results of the ablation study on the benchmark are presented in \cref{tab:results_ablation_benchmark}. As shown in Tab~\ref{tab:results_ablation_benchmark}, significant improvements in the consistency metrics (DC and D\&C-DS) are observed after individually adding RA and IA. The best performance is achieved when incorporating both components.
In the 3-subject scenario, the improvement from adding RA alone is minimal, but it becomes more significant when combined with IA. This is likely due to the stronger internal attraction between multiple subjects in the 3-subject case. Without isolating these attractions, repositioned features cannot be fully utilized.
These results demonstrate the effectiveness of our proposed modules.

\subsubsection{Subjective Comparison}

\cref{tab:userstudy_abl} presents the results from our user study conducted for the ablation study. The results demonstrate that our method surpasses all ablation models in consistency metrics, which is our primary objective.
\cref{fig:ablation_study} shows images generated by incrementally incorporating individual components (RA and IA). When only IA is applied, our method effectively prevents the merging of multiple subjects, as observed in the comparison of the dog in the third row with that in the second row, first column.
Adding RA further enhances detail consistency (e.g., the dog in the first column)  by more effectively leveraging repositioned reference features. These results confirm the effectiveness of our proposed modules.


\section{Conclusion} \label{sec:conclusion}

In this paper, we propose IR-Diffusion, a training-free diffusion model to enhance multi-subject consistency. We first identify two key issues in existing diffusion models: the positional impact of the attention mechanism and the undesired internal attraction among multiple subjects. To address these challenges, we introduce Isolation Attention, which prevents subject convergence caused by internal mutual attraction, and Reposition Attention, which aligns subjects in reference and target images to the same positions. Our method demonstrates significant performance improvements in consistency metrics, surpassing all existing methods in open-domain scenarios. Additionally, these findings contribute to advancing the field by revealing the underlying mechanisms of diffusion models. We believe these findings also hold broader potential for applications such as attribute-binding and video generation. 
In future work, we will extend our approach to the DiT architecture, e.g., FLUX~\cite{flux2023}, and PixArt-$\Sigma$~\cite{chen2023pixart}.

{\small
\bibliographystyle{ieeenat_fullname}
\bibliography{ref}
}

\clearpage
\appendix
\section*{Supplementary Material}

This supplementary material provides an extensive review of related works on story visualization, which leverages large story image datasets to enable models to generate sequential images, as detailed in Sec.~\ref{supp:related_sv}. The Isolation Attention and Reposition Attention mechanisms in our method are implemented using a masking mechanism, followed by previous works~\cite{cao2023masactrl, tewel2024_ConsiStory, storydiffusion, dreamstory}. Detailed implementation can be found in Sec.~\ref{supp:mask_imple}.
Finally, additional results and discussions are presented in Sec.~\ref{sec:more_results}.

\section{Related Works of Story Visualization}\label{supp:related_sv}


Early story visualization methods~\cite{SV_GAN1, SV_GAN2, SV_T1, CVPR2023_Make_a_Story, gu2023tevis_retrieval, storydalle, ARLDM} initially relied on curated datasets, such as PororoSV~\cite{StoryGAN_2019_CVPR} and FlintstonesSV~\cite{FlintstonesSV}.
For instance, Rahman et al.~\cite{CVPR2023_Make_a_Story} proposed an innovative autoregressive diffusion-based framework, including a visual memory module that implicitly captures actor and background context across the generated frames. Similarly, Pan et al.~\cite{ARLDM} proposed an autoregressive diffusion model conditioned on historical captions and generated images, utilizing multimodal guidance from a CLIP~\cite{radford2021CLIP} text encoder and a BLIP~\cite{BLIP, BLIP2} multimodal encoder to ensure coherent and relevant image generation. Further, Liu et al.~\cite{storygen_liu2023intelligent} introduced the StorySalon dataset and achieved state-of-the-art results.

However, these methods are significantly constrained by the limitations of existing datasets in terms of size and quality, which hampers their performance in open-domain tasks. In contrast, our training-free approach does not require additional datasets. By leveraging foundation models trained on extensive datasets (e.g., LAION-5B~\cite{schuhmann2022laion_5B}), it is particularly well-suited for open-domain scenarios.

\section{Implementation Details}\label{supp:mask_imple}

\subsection{Existing Attention Mechanism}\label{supp:attn}

In popular diffusion models (e.g., SD~\cite{sd_ldm_diffusion}, and SD-XL~\cite{podell2023sdxl}), the attention mechanism within the U-Net network typically consists of a self-attention layer followed by a cross-attention layer. 

For clarity, the residual connections and layer count are omitted, and a standard attention layer can be formulated as follows, 
\begin{align}
\label{supp_eq_softmax}
    O_i &= \textit{softmax}\left({Q_i K_i}\right) \cdot V_i,
\end{align}
where $A_i$ represents the attention weights, and $O_i$ denotes the output of the attention layer for the \textit{i}-th image. Here, $Q$ refers to the query features derived from spatial features, while $K$ and $V$ represent the Key and Value features. These $K$ and $V$ are obtained from spatial features in self-attention layers with specific projection matrices.

Recent studies have shown that appearance information can be incorporated into the generation process by cascading~\cite{tewel2024_ConsiStory, storydiffusion, dreamstory} the $K$ and $V$ features from those references. 
We use the subscript ‘TGT’ to denote the features of the target image and the subscript ‘REF’ to represent the features from the reference image. Therefore, their attention can be calculated as follows,
\begin{align}
    K^{+} &= [K^{1}_\text{\scriptsize \tiny REF} \oplus K^{2}_\text{\scriptsize \tiny REF} \oplus \ldots \oplus K^{N}_\text{\scriptsize \tiny REF} \oplus K_\text{\scriptsize \tiny TGT}],  \\
    V^{+} &= [V^{1}_\text{\scriptsize \tiny REF} \oplus V^{2}_\text{\scriptsize \tiny REF} \oplus \ldots \oplus V^{N}_\text{\scriptsize \tiny REF} \oplus V_\text{\scriptsize \tiny TGT}],  \\
    M^{+} &= [M^1 \oplus M^2 \oplus \ldots \oplus M^{N} \oplus \mathds{1}],\label{eq:mask} \\
    O_\text{\scriptsize \tiny TGT} &= \textit{softmax}\left({Q_\text{\scriptsize \tiny TGT} K^{+}} + \log M^{+} \right) \cdot V^{+},
\end{align}
where $M_{i}$ is the subject mask for \textit{i}-th subject, and $\oplus$ indicates the concatenation operation.

However, these approaches failed to consider the intrinsic properties of the attention mechanism inherent to the diffusion model, i.e., multi-subject internal attraction and influence of position.

To this end, we proposed Reposition Attention and Isolation Attention. Specifically, Isolation Attention focuses on eliminating internal attraction between different subjects. On the other hand, Reposition Attention aims to rescale and relocate the features to the optimal positions, ensuring the information in the attention mechanism can be effectively utilized.

\begin{table*}[t]
\setlength{\abovecaptionskip}{-0.1cm} 
    \centering
    \small \renewcommand{\arraystretch}{1.25}
    \caption{Quantitative results of different backbone for our DreamStory on the DS-500 benchmark. In each backbone, the best consistency metrics (DC and D\&C-DS) are highlighted in bold.
    Our IR-Diffusion method has achieved significant improvements in consistency across various backbones, demonstrating its generalizability and robustness.
    }
    \label{supp_tab:results_ablation_backbone}\

\begin{tabular}{p{4.2cm}| c c c c | c c c c}
\bottomrule[1.5pt]
\multicolumn{1}{l|}{\multirow{2}{*}{}}                               & \multicolumn{4}{c|}{\textbf{2-Subject}} & \multicolumn{4}{c}{\textbf{3-Subject}} \\
\multicolumn{1}{l|}{}                                               & AES$\uparrow$ & CLIP-T$\uparrow$  & DS$\uparrow$  &  D\&C-DS$\uparrow$   & AES$\uparrow$ & CLIP-T$\uparrow$  & DS$\uparrow$  &  D\&C-DS$\uparrow$   \\ \hline
SDXL~\cite{podell2023sdxl}         & 6.53    & 0.3823   & 0.4949   &  0.3116    & 6.56   & 0.3994 & 0.4548   & 0.1506 \\
SDXL~\cite{podell2023sdxl} + \textbf{Ours}  & 6.58 & 0.3729 & \textbf{0.6224} & \textbf{0.4322} & 6.65 & 0.3816 & \textbf{0.5925} & \textbf{0.3210} \\ 

\hline \hline
Playground~\cite{li2024playground}         & 6.66    & 0.3782   & 0.5773   & 0.4195  & 6.80   & 0.3884   & 0.5022   & 0.2026   \\
Playground~\cite{li2024playground} + \textbf{Ours} & 6.64 & 0.3679   & \textbf{0.7518}  & \textbf{0.6458} & 6.68 & 0.3736   & \textbf{0.6742} & \textbf{0.4095}  \\ 

\hline \hline
Kolors~\cite{kolors}         & 6.49    & 0.3744   & 0.5280   & 0.3776  & 6.55   & 0.3794   & 0.4966   & 0.1918   \\
Kolors~\cite{kolors} + \textbf{Ours} & 6.45 & 0.3612   & \textbf{0.6834} & \textbf{0.5293} & 6.48 & 0.3681   & \textbf{0.6389} & \textbf{0.3147}  \\
\toprule[1.5pt]
\end{tabular}
\vspace{-0.5cm}
\end{table*}

\subsection{Isolation Attention}

Vanilla self-attention computes responses between every pair of tokens within the same image, resulting in the attention map. 
However, as discussed in the main paper, different subjects are attracted to each other in the self-attention layer. This attration may lead to subject fusion, where multiple subjects merge into a single entity, simultaneously exhibiting characteristics of both original subjects.
To mitigate this attraction, each subject should not affect others during the vanilla forward process of the target image. We achieve this by masking mechanisms to ensure that each subject does not receive responses from the other subjects' $k$$v$. Specifically, we apply masks to zero out the undesired attention map responses between different subjects.

Assuming the masks for \textit{i}-th subjects in the target image are $m_{i}$, the final isolation attention mask for the target image ($M_\text{\scriptsize \tiny TGT}$) can be obtained by iteratively multiplying the complement of the \textit{j}-th mask with the \textit{i}-th mask. Therefore, the $M_\text{\scriptsize \tiny TGT}$ can be calculated as follows,
\begin{align}
    M_\text{\scriptsize \tiny TGT} &= \prod_{}^{} \left [  F(m_i) \times F(1-m_j)^T \right ] ,   & i \ne j.\label{eq:mask_tgt}
\end{align}
The $F(\cdot)$ is a flattened operation, and the symbol $\prod$ indicates multiplication. Followed by previous works~\cite{cao2023masactrl, tewel2024_ConsiStory, dreamstory}, the standard attention masking is adopted, which nullifies softmax's logits by assigning corresponding scores to $-\infty$. By replacing the all-ones mask ($\mathds{1}$) in \cref{eq:mask} with the above $M_\text{\scriptsize \tiny TGT}$, we can prevent internal attraction between different subjects by ensuring that each subject’s $Q$ does not receive responses from the $K$$V$ in other subject’ regions. So the final $M^{+}$ in the \cref{eq:mask} can be re-formulated as follows, 
\begin{align}
    M^{+} &= [M_1 \oplus M_2 \oplus \ldots \oplus M_{N} \oplus M_\text{\scriptsize \tiny TGT}],\label{eq:mask_final}
\end{align}
It is essential to highlight that the key distinction between our proposed IA and the existing DreamStory method is the substitution of the all-ones matrix in \cref{eq:mask} with our novel isolation matrix, $M_\text{\scriptsize \tiny TGT}$ in \cref{eq:mask_tgt}. This isolation matrix isolates internal multi-subject attractions, enabling independent subject generation and preventing the convergence of multiple subjects into one single entity.
This adjustment ensures that each subject's unique attributes are preserved, leading to more accurate and coherent results in the overall process.

\subsection{Reposition Attention}

As demonstrated in the main paper, self-attention tends to reference tokens in neighboring positions. Our Reposition Attention (RA) is designed to rescale and reposition the features of the reference subject to match the positions of the corresponding subjects in the target image. 

Assuming the repositioned $KV$ features are denoted with a hat ($\hat{K}\hat{V}$), the formulation of our Reposition Self-Attention is as follows,
\begin{align}
    \hat{K}^{+} &= [\hat{K}^{1}_\text{\scriptsize \tiny REF} \oplus \hat{K}^{2}_\text{\scriptsize \tiny REF} \oplus \ldots \oplus \hat{K}^{N}_\text{\scriptsize \tiny REF} \oplus K_\text{\scriptsize \tiny TGT}],  \\
    \hat{V}^{+} &= [\hat{V}^{1}_\text{\scriptsize \tiny REF} \oplus \hat{V}^{2}_\text{\scriptsize \tiny REF} \oplus \ldots \oplus \hat{V}^{N}_\text{\scriptsize \tiny REF} \oplus V_\text{\scriptsize \tiny TGT}],  \\
    O_\text{\scriptsize \tiny TGT} &= \textit{softmax}\left({Q_\text{\scriptsize \tiny TGT} \hat{K}^{+}} + \log M^{+} \right) \cdot \hat{V}^{+}, 
\end{align}
The $\hat{K}_{i}$ and $\hat{V}_{i}$ features will be rescaled and relocated before projection. Thus, the visual features from the reference images are relocated to corresponding subjects' positions in target images, enhancing the utilization of these features.

\subsection{More Implementation Details} \label{supp:more_implements_details}

A guidance scale of 7.0~\cite{ho2021classifier} and the default scheduler with 50 inference steps~\cite{lu2022dpmnew} are used. 
All methods are evaluated on their respective generated images with a resolution of $1280\times768$ pixels.
We also applied Masked Mutual Self-Attention (MMSA) and Masked Mutual Cross-Attention (MMCA) from DreamStory~\cite{dreamstory} for multi-subject consistent generation. Additionally, a 0.5 token dropout rate was used to increase the diversity of generated subject poses, followed by previous works~\cite{tewel2024_ConsiStory, storydiffusion, dreamstory}. To prevent inaccuracies in the number of generated masks, we set a maximum of 20 attempts for retry. This engineering trick averages 4.97 attempts for the 3-subject benchmark, accounting for approximately 15 seconds (22\% of the total time).  It is simple yet effective, ensuring accurate initial subject mask generation for fairer comparisons.

\section{Experiments}\label{sec:more_results}

\subsection{Consistency Across Various Backbones}

To validate the generalizability and robustness of our method, we integrated it into three popular backbones, i.e., SD-XL~\cite{podell2023sdxl}, Playground~\cite{li2024playground}, and Kolors~\cite{kolors}. All the results are presented in \cref{supp_tab:results_ablation_backbone}.
As shown in \cref{supp_tab:results_ablation_backbone}, significant improvements were observed across different backbones after incorporating our method. Among the tested backbones, Playground exhibited the highest performance, surpassing the others in both aesthetic and consistency metrics.

\subsection{More Generation Results}

\subsubsection{More Results vs. SOTA Methods}

In this subsection, we provide additional results of IR-Diffusion and other SOTA methods in \cref{fig:supp_main_results_1} and \cref{fig:supp_main_results_2}. These results demonstrate that our IR-Diffusion achieves superior multi-subject consistency performance compared to other SOTA methods in open-domain scenarios.

In \cref{fig:supp_vs_MSDiffusion_1}, we present a comparison of 3-subject anime style generation between our IR-Diffusion and MS-Diffusion~\cite{MSDiffusion}. 
Additionally, \cref{fig:supp_vs_MS_SM_1} presents comparisons involving both MS-Diffusion and StoryMaker~\cite{storymaker} in 2-subject scenarios. StoryMaker is excluded from the 3-subject comparisons in \cref{fig:supp_vs_MSDiffusion_1} due to its dataset limitation, which supports only the synthesis of one or two human faces.
Our method demonstrates superior performance, producing more visually appealing and consistent results. Unlike MS-Diffusion and StoryMaker, which are constrained by their limited datasets (primarily featuring general subject photos), our approach excels in open-domain scenarios and consistently generates high-quality anime-style outputs.

We also present more visual results of our IR-Diffusion with different styles in \cref{fig:supp_ours_1}. These results show that our approach can consistently generate high-quality images in open-domain scenarios.

\subsubsection{More Results for Ablation Study}
More visual results of our ablation study are presented in \cref{fig:supp_ABL_1} and \cref{fig:supp_ABL_heatmap_1}.
\cref{fig:supp_ABL_1} illustrates generation results from different ablation configurations. The incorporation of Reposition Attention (RA) and Isolation Attention (IA) individually led to notable improvements in the consistency of the generated outputs. When both mechanisms were applied, our IR-Diffusion achieved superior multi-subject consistency, demonstrating the effectiveness of these attention mechanisms.

\cref{fig:supp_ABL_heatmap_1} provides mutual attention heatmap visualizations, revealing significantly higher attention values for reference image features with IR-Diffusion. This highlights the improved utilization of reference image information and demonstrates how the RA mechanism enhances consistency by aligning subjects in the reference and target images.

These results all prove the effectiveness of RA and IA in achieving superior consistency and quality in open-domain image generation.

\subsubsection{Efficiency Analysis}

We studied the runtime overhead and conducted our tests using an H800 GPU. 
The runtime cost of our IR-Diffusion and other methods are presented in \cref{tab:time}. Compared to the baseline (DreamStory), IA and RA all introduce approximately 5s and 8s of additional computation time for 2-Subject and 3-Subject scenarios, respectively. When applied together, the overhead increases to approximately 10s~(45\%) and 16s~(57\%), yet the total processing time remains within 1 minute, which is acceptable. In contrast, finetuning-based approaches such as MuDI require approximately 1.5 hours and 2 hours of finetuning per case for 2-Subject and 3-Subject scenarios, respectively. This highlights the efficiency advantage of our method.

\begin{table}[t]
\setlength{\abovecaptionskip}{0.01cm} 
\setlength{\belowcaptionskip}{-0.1cm}  
    \centering
    \small \renewcommand{\arraystretch}{1.25}
    \caption{Average time Consumption on different methods. 
    It includes the time for generating reference subjects and the final scene. Additionally, MUDI requires extra fine-tuning time for each case.}
    \label{tab:time}
\begin{tabular}{p{3cm} p{2cm}<{\centering} p{2cm}<{\centering}}
\bottomrule[1.0pt]
                 & 2-Subject & 3-Subject \\
\hline
MuDI~\cite{MUDI}                        & 5400s & 7200s \\
MS-Diffusion~\cite{MSDiffusion}         &  10s  &  13s  \\
ConsiStory~\cite{tewel2024_ConsiStory}  &  30s  &  38s  \\
StoryDiffusion~\cite{storydiffusion}    &  21s  &  25s  \\
DreamStory~\cite{dreamstory}            &  22s  &  28s  \\
\textbf{IR-Diffusion} (\textbf{Ours})   &  32s  &  44s  \\
\bottomrule[1.0pt]
\end{tabular}
\vspace{-5pt}
\end{table}

\subsection{Limitations and Failure Cases} \label{subsec:exp_limitation}

\begin{figure}[t]
    \centering
    \includegraphics[width=1.0\columnwidth]{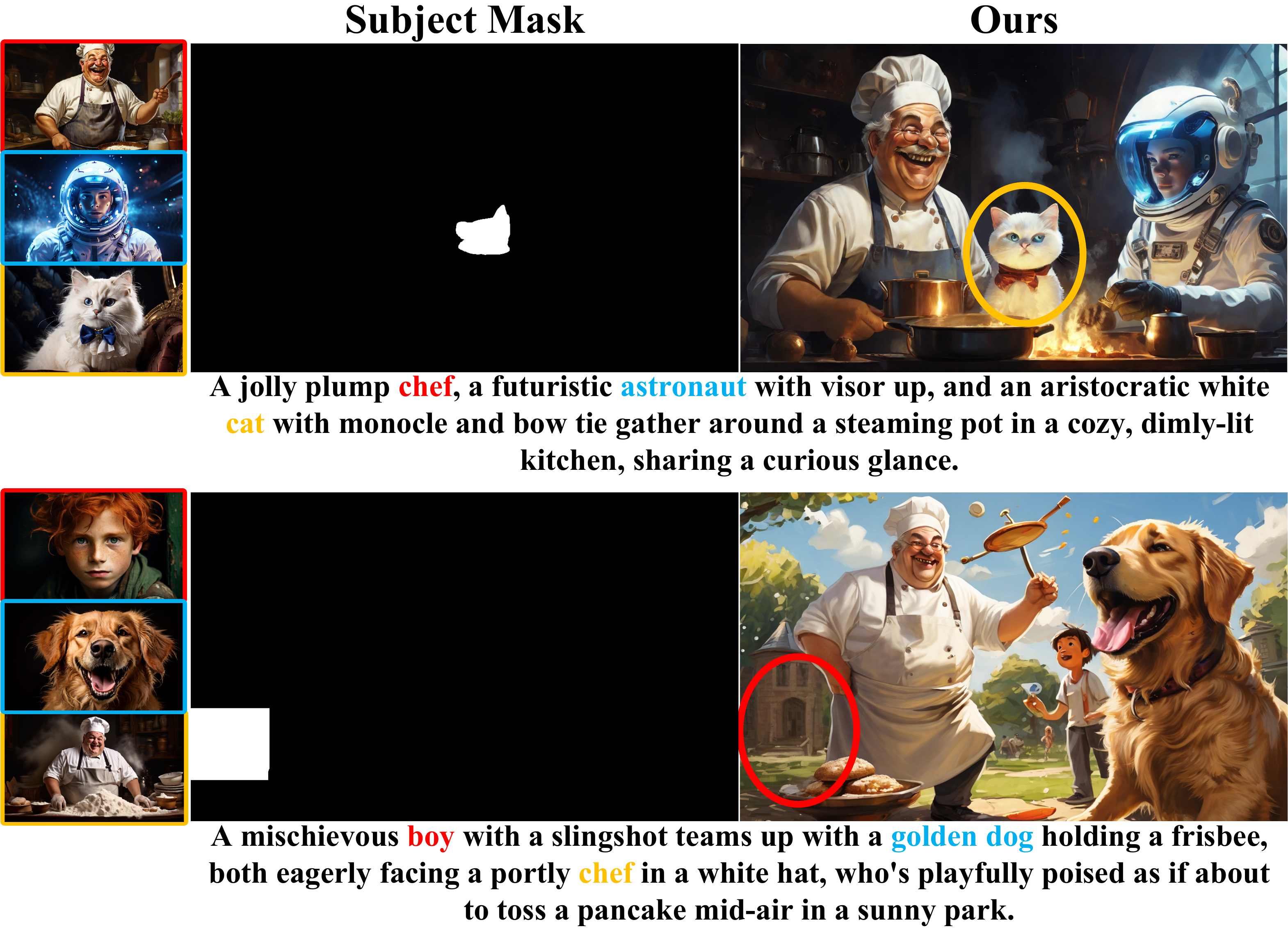}
    \caption{Failure cases in scenarios with small subject masks.}
    \label{fig:failure_case}
    \vspace{-10pt}
\end{figure}

Our method may experience performance degradation when the given mask is very small. This common limitation in masked-based methods is mainly due to the downsampling process in VAE~\cite{VAE} ($\downarrow\times8$) and U-Net~\cite{ronneberger2015unet} ($\downarrow\times4$), where small regions are compressed into just a few tokens, limiting the model's ability to generate fine-grained details (cat in the first row of \cref{fig:failure_case}), or even leading to target missing (boy in the second row of \cref{fig:failure_case}).

\begin{figure*}[t]
    \centering
    \setlength{\abovecaptionskip}{0.5cm}
    \includegraphics[width=1.0\textwidth]{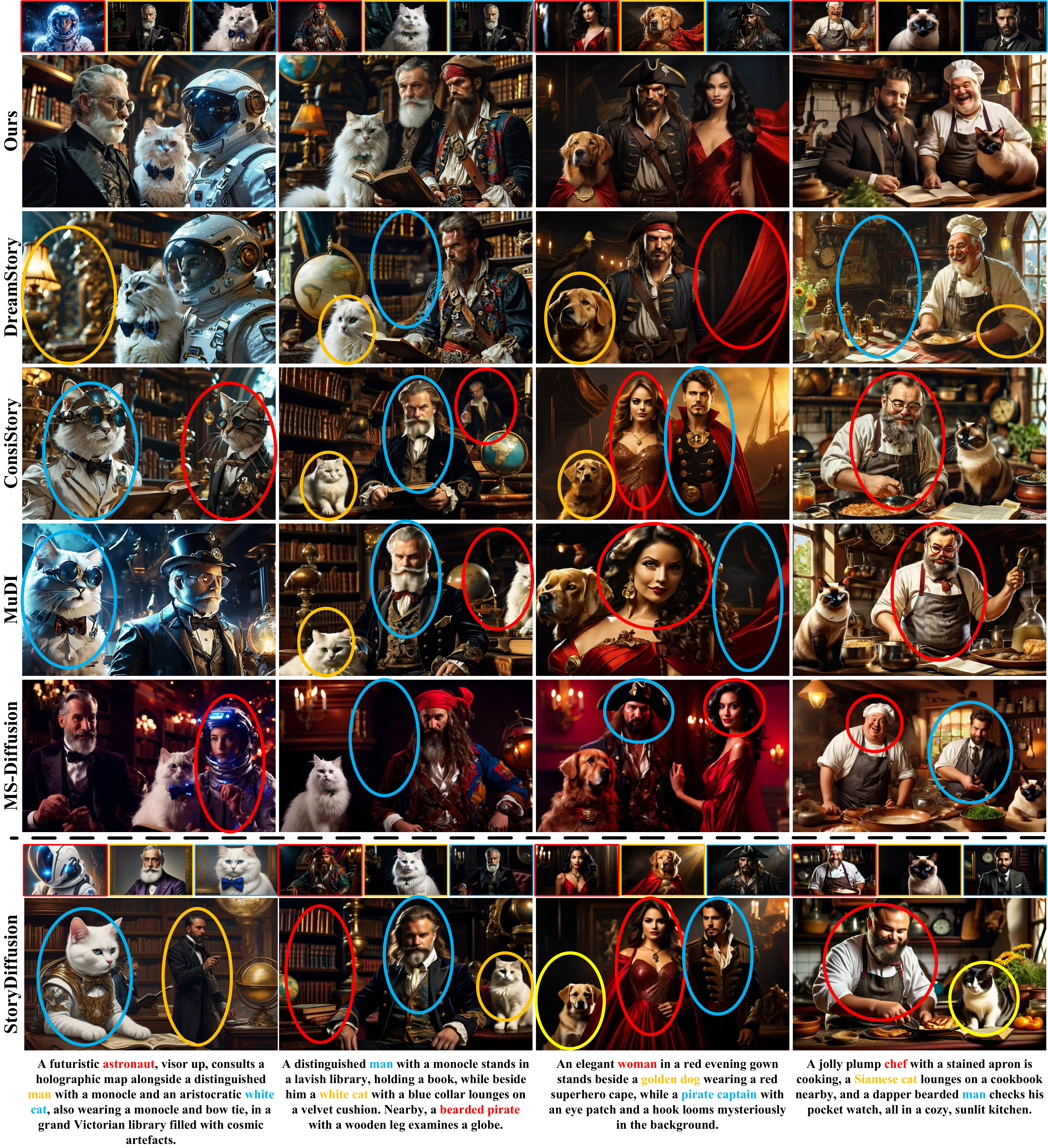}
    \caption{Comparisons of multi-subject consistency generation between our IR-Diffusion and other SOTA methods. The superior performance of our approach is evident from the more visually appealing and consistent results. Except for StoryDiffusion, which uses the portraits above it for reference, all other methods use the top portraits as a reference. Different subjects are indicated with different colors. }
    \label{fig:supp_main_results_1}
    \vspace{0.5cm}
\end{figure*}

\begin{figure*}[t]
    \centering
    \setlength{\abovecaptionskip}{0.5cm}
    \includegraphics[width=1.0\textwidth]{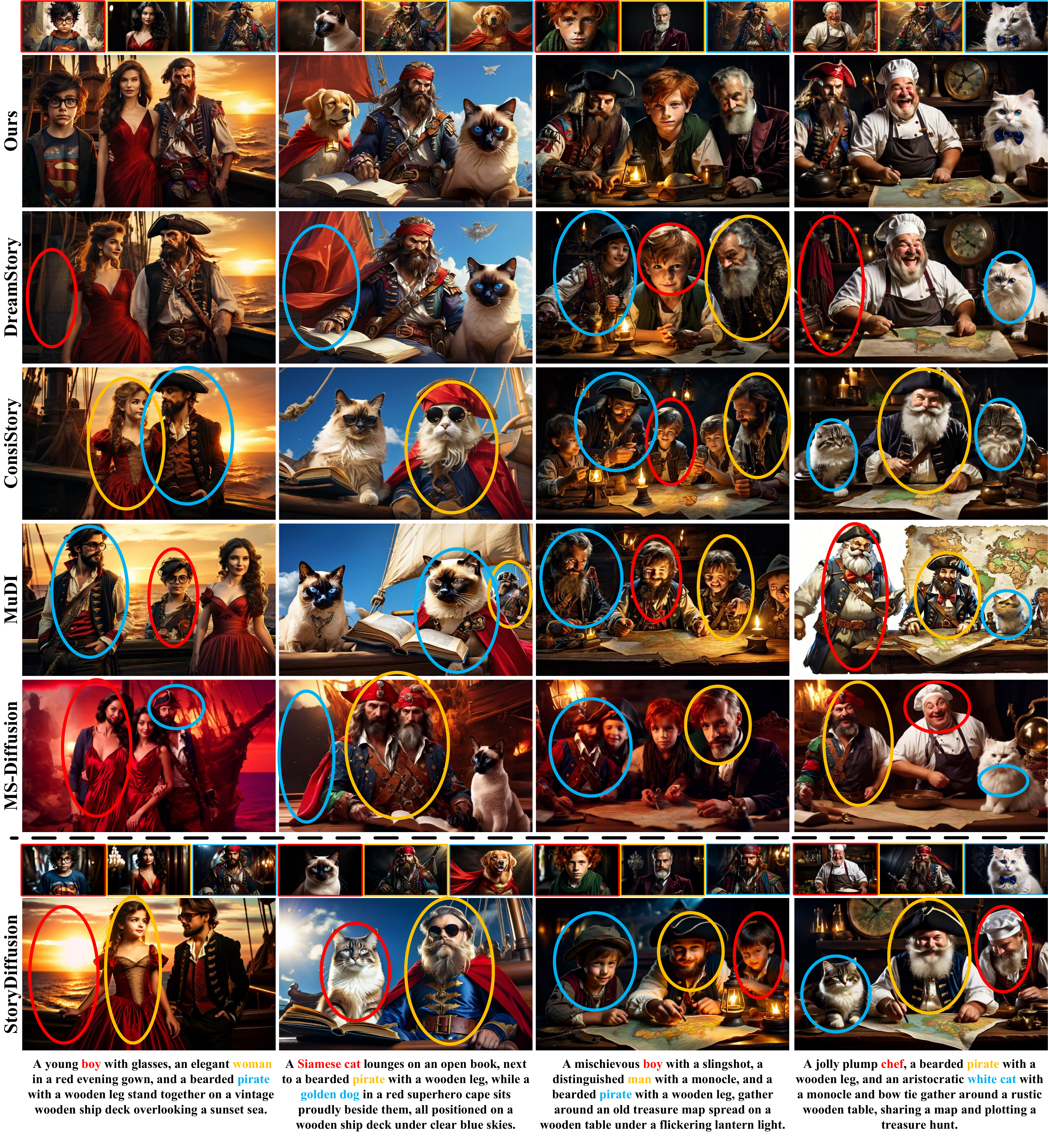}
    \caption{Comparisons of multi-subject consistency generation between our IR-Diffusion and other SOTA methods. The superior performance of our approach is evident from the more visually appealing and consistent results. Except for StoryDiffusion, which uses the portraits above it for reference, all other methods use the top portraits as a reference. Different subjects are indicated with different colors. }
    \label{fig:supp_main_results_2}
    \vspace{0.5cm}
\end{figure*}

\begin{figure*}[t]
    \centering
    \setlength{\abovecaptionskip}{0.5cm}
    \includegraphics[width=1.0\textwidth]{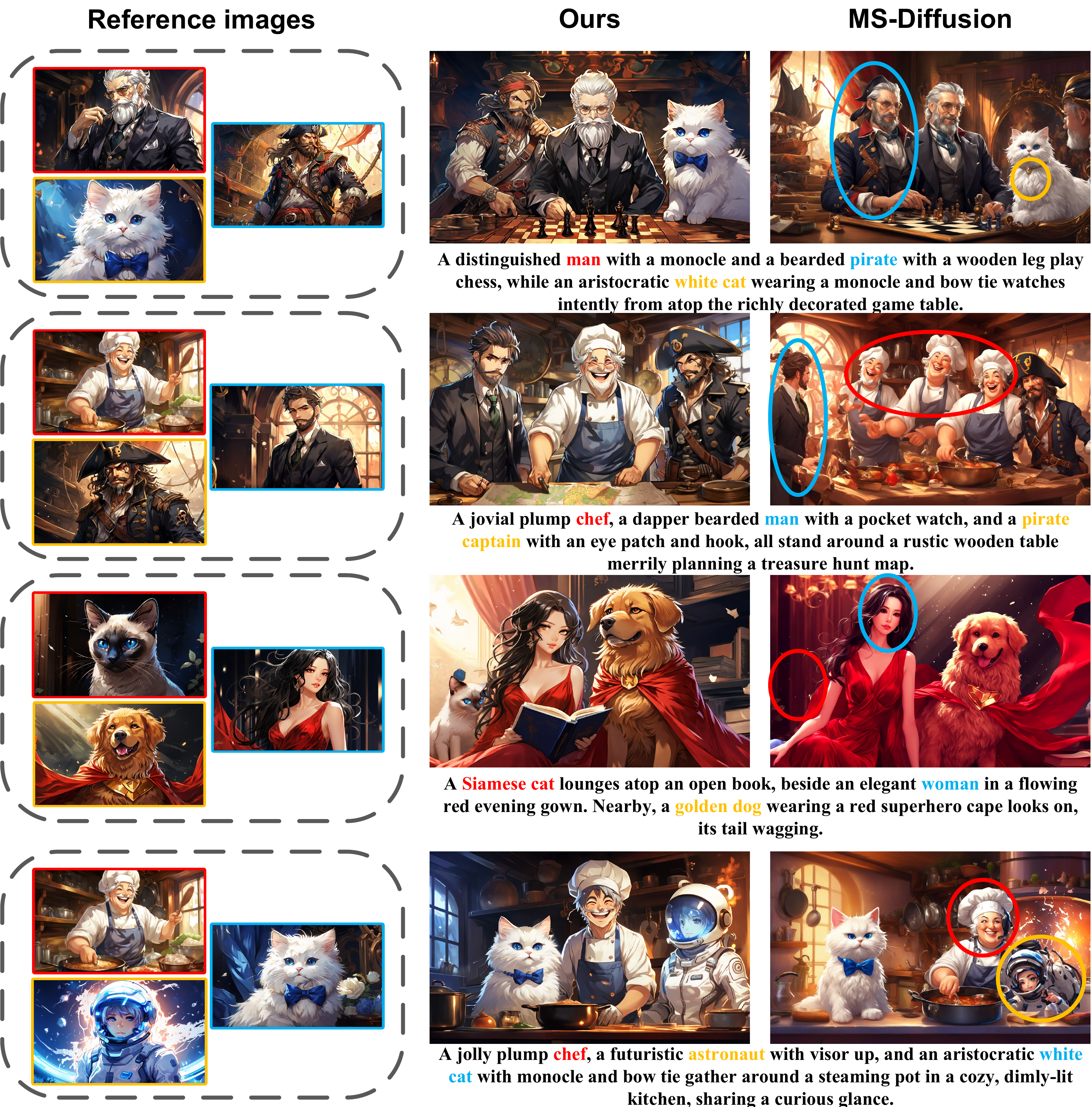}
    \caption{Comparisons of anime style generation between our IR-Diffusion and MS-Diffusion~\cite{MSDiffusion}.
    The superior performance of our approach is evident from the more visually appealing and consistent results in the anime-style generation. While MS-Diffusion is limited by its dataset and unsuitable for open-domain applications, our method excels in generating consistent and high-quality anime styles. Different subjects are indicated with different colors.
    }
    \label{fig:supp_vs_MSDiffusion_1}
    \vspace{0.5cm}
\end{figure*}

\begin{figure*}[t]
    \centering
    \setlength{\abovecaptionskip}{0.5cm}
    \includegraphics[width=1.0\textwidth]{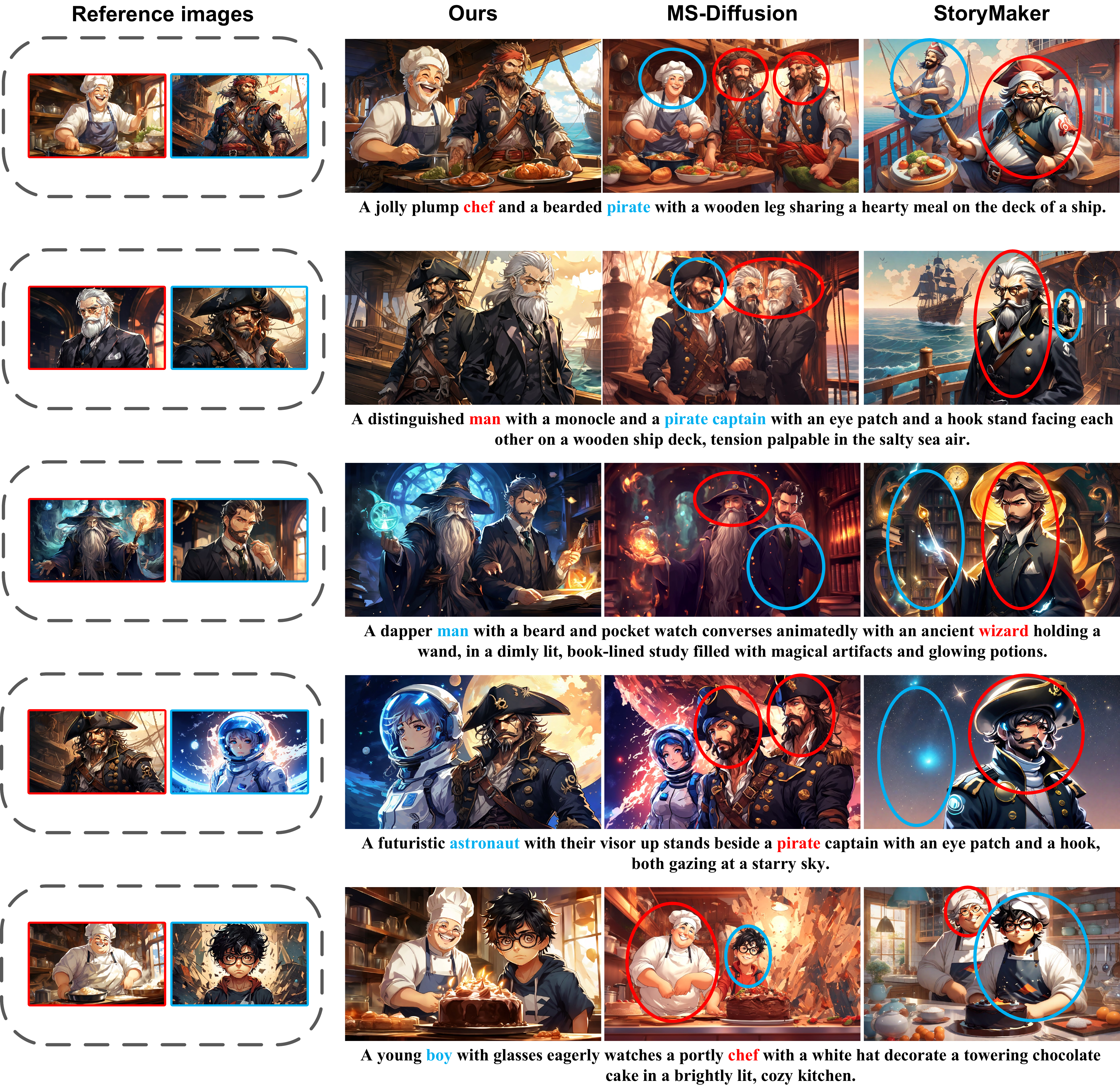}
    \caption{Comparisons of anime style generation between our IR-Diffusion, MS-Diffusion~\cite{MSDiffusion} and StoryMaker~\cite{storymaker}.
    The superior performance of our approach is evident from the more visually appealing and consistent results in the anime-style generation. While MS-Diffusion and StoryMaker are limited by their datasets and unsuitable for open-domain applications, our method excels in generating consistent and high-quality anime styles. Different subjects are indicated with different colors.
    }
    \label{fig:supp_vs_MS_SM_1}
    \vspace{0.5cm}
\end{figure*}

\begin{figure*}[t]
    \centering
    \setlength{\abovecaptionskip}{0.5cm}
    \includegraphics[width=1.0\textwidth]{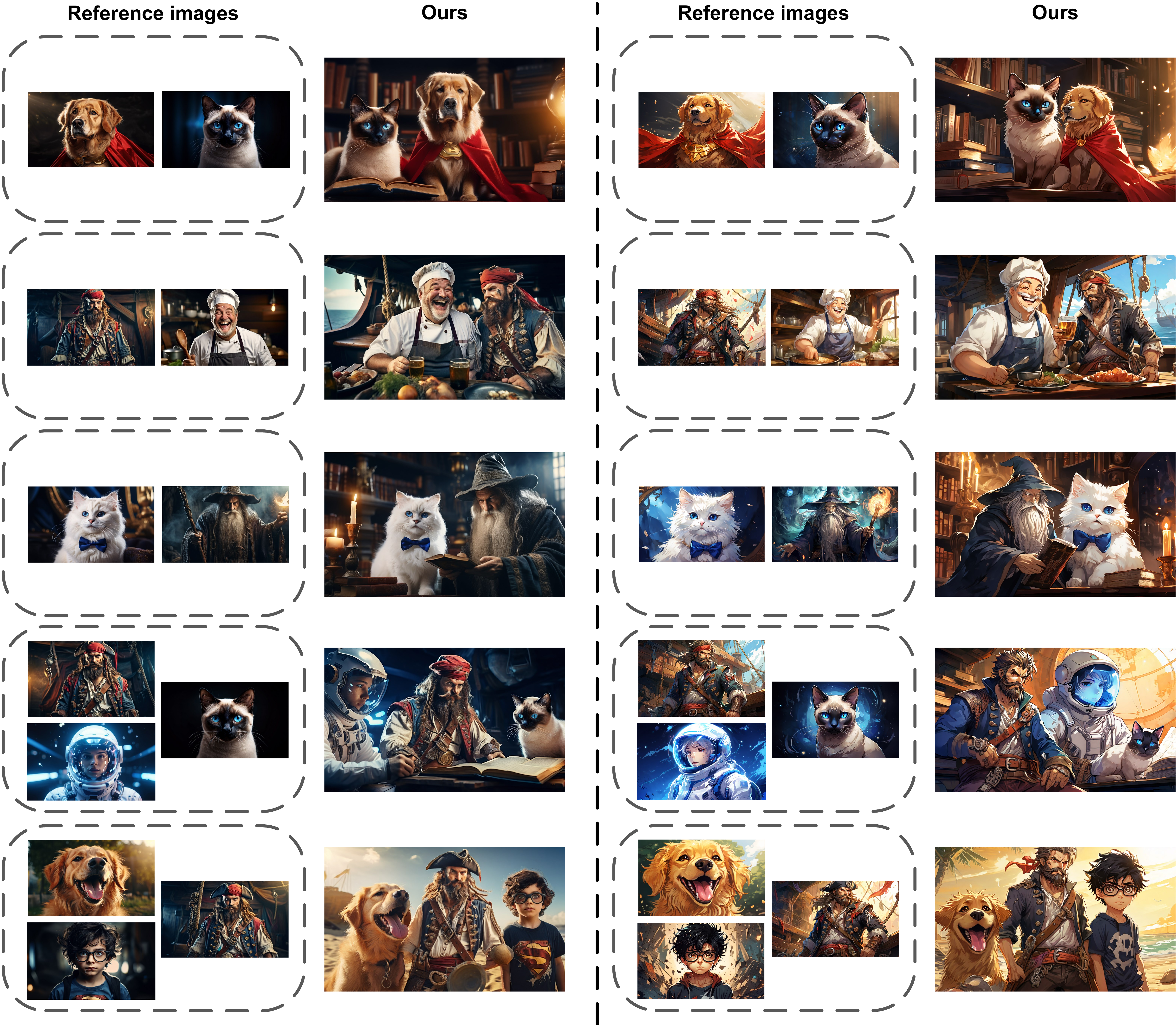}
    \caption{Generated images from our IR-Diffusion. 
    Real-style images are displayed on the left, and anime-style images on the right. Each row corresponds to the same case across different styles. The results demonstrate the superior performance and versatility of our approach in generating high-quality images in open-domain scenarios.
    }
    \label{fig:supp_ours_1}
    \vspace{0.5cm}
\end{figure*}

\begin{figure*}[t]
    \centering
    \setlength{\abovecaptionskip}{0.5cm}
    \includegraphics[width=1.0\textwidth]{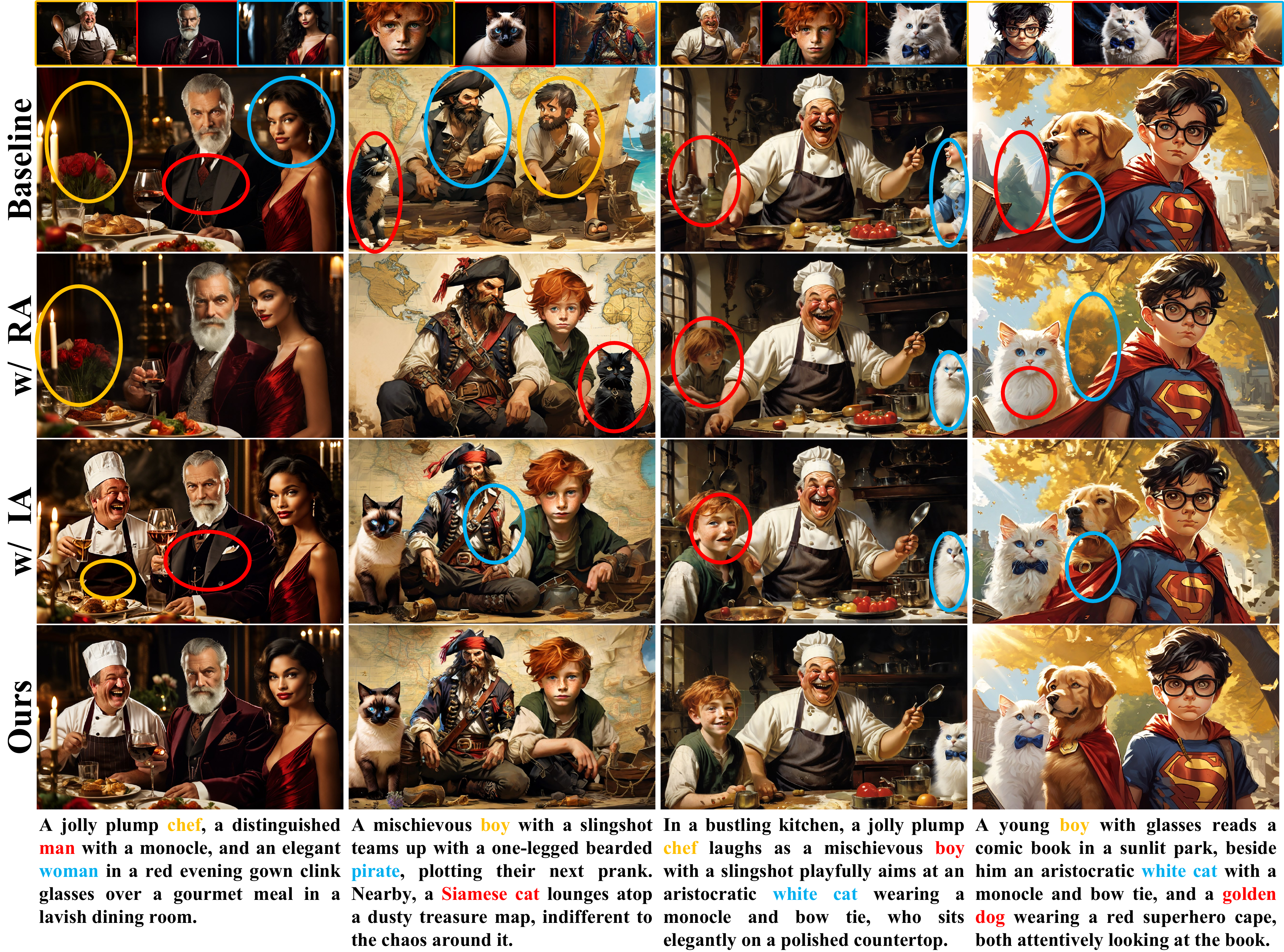}
    \caption{Ablation studies of different generation results.
    All methods use the top portraits as a reference. Different subjects are indicated with different colors. 
    The consistency of the generated outputs improved significantly with the individual incorporation of Reposition Attention (RA) and Isolation Attention (IA). When both RA and IA were applied, our IR-Diffusion achieved superior multi-subject consistency, demonstrating the effectiveness of IA and RA.
    }
    \label{fig:supp_ABL_1}
    \vspace{0.5cm}
\end{figure*}

\begin{figure*}[t]
    \centering
    \setlength{\abovecaptionskip}{0.5cm}
    \includegraphics[width=0.8\textwidth]{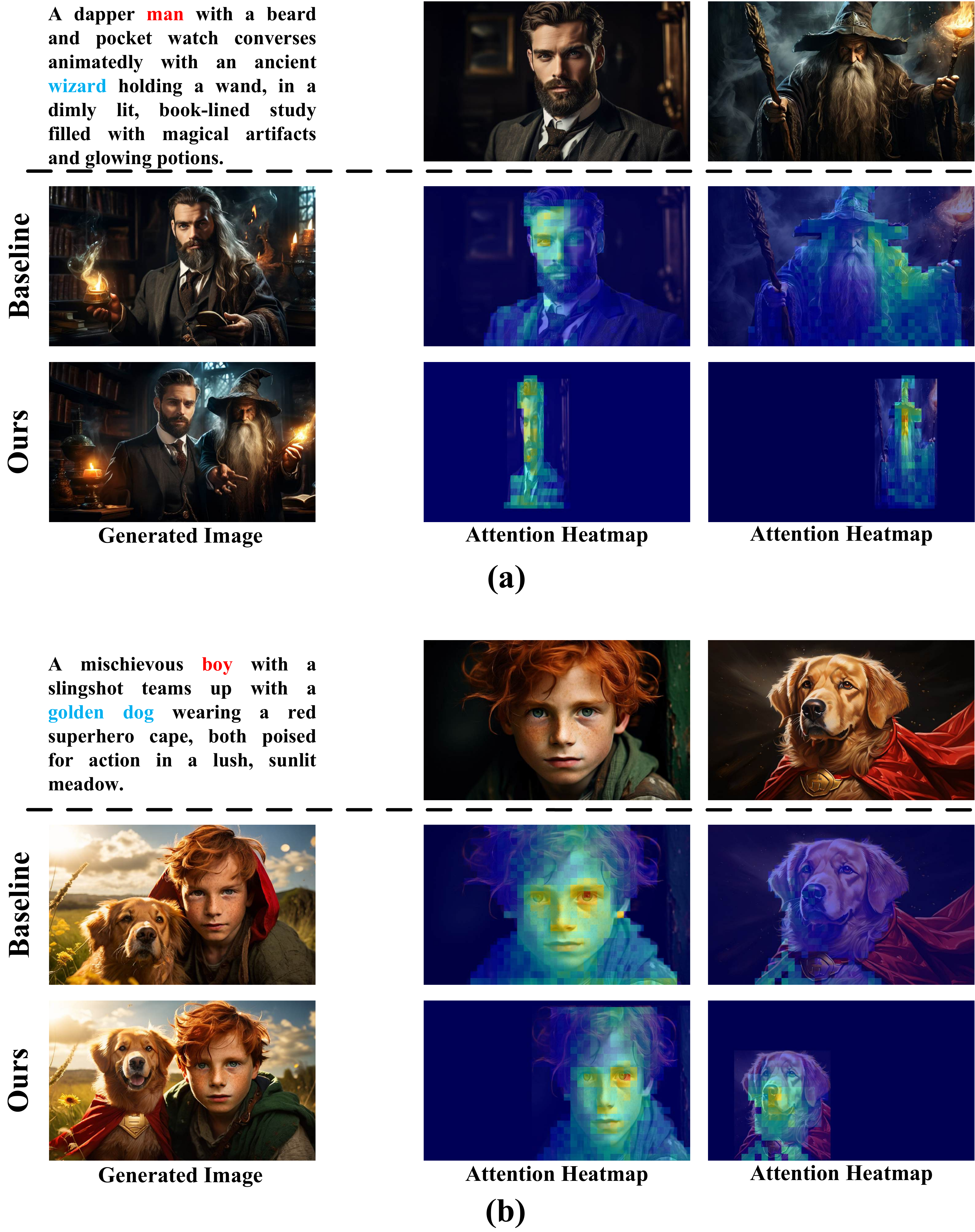}
    \caption{Mutual attention heatmap visualization for the ablation study.
        With the integration of our IR-Diffusion, reference image features exhibit significantly higher attention values, demonstrating enhanced utilization of reference image information. These results highlight the effectiveness of the Reposition Attention (RA) mechanism in improving consistency by aligning subjects between reference and target images.
    }
    \label{fig:supp_ABL_heatmap_1}
    \vspace{0.5cm}
\end{figure*}

\end{document}